\documentclass[a4paper,twocolumn]{article}
\usepackage{hyperref}
\usepackage{graphicx}
\usepackage{amsmath}
\usepackage{amsfonts}
\usepackage[english]{babel}
\usepackage{amsmath}
\usepackage{xcolor}
\usepackage{wasysym}
\usepackage{hyperref}
\usepackage{amsthm}
\usepackage{simpleconf}
\usepackage{floatpag}

\usepackage{geometry}

\usepackage[%
  backend=biber,
  style=ieee, %
  hyperref=auto,
  doi=false,
  isbn=false,
  url=false,
  eprint=false,
  maxbibnames=13,
  ]{biblatex}
\usepackage{subfig}

\newcommand{\bs}[1]{\boldsymbol{#1}}

\newcommand{\qq}{\bs{q}}
\newcommand{\dqq}{\dot{\bs{q}}}

\newcommand{\ddqq}{\ddot{\bs{q}}}

\newcommand{\dq}{\dot{q}}

\newcommand{\gggg}{\bs{g}}

\newcommand{\pp}{\bs{p}}

\newcommand{\CC}{\bs{C}}

\newcommand{\MM}{\bs{M}}

\newcommand{\ggamma}{\bs{\gamma}}

\newcommand{\nnull}{\bs{0}}

\renewcommand{\d}{{\mathrm{d}}}

\makeatletter
\renewcommand*\env@matrix[1][\arraystretch]{%
	\edef\arraystretch{#1}%
	\hskip -\arraycolsep
	\let\@ifnextchar\new@ifnextchar
	\array{*\c@MaxMatrixCols c}}
\makeatother

\definecolor{TUMBlue}{HTML}{0065BD}

\newcommand{\q}{\boldsymbol{q}}
\newcommand{\qdot}{\dot{\boldsymbol{q}}}

\DeclareUnicodeCharacter{2212}{-}

\AtBeginBibliography{\small}

\newtheorem{proposition}{Proposition}
\newtheorem{conjecture}{Conjecture}
\usepackage{nicefrac}
\usepackage{url}
\usepackage{blindtext}

\graphicspath{{pdftex}{figures}}
\bibliography{bibliography}

\author{
	\begin{tabular}{cc}
		\begin{minipage}{.45\textwidth}
			\normalsize
			\begin{center}
	\href{https://orcid.org/0000-0001-5343-9074}{\includegraphics[scale=0.06]{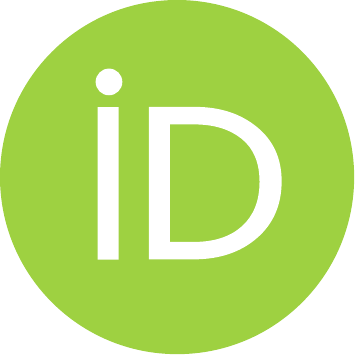}\hspace{1mm}\textbf{Alin Albu-Schäffer}} \\
	Institute of Robotics and Mechatronics\\
	German Aerospace Center (DLR)\\
	Oberpfaffenhofen, Germany\\
	\texttt{alin.albu-schaeffer@dlr.de}
			\end{center}
		\end{minipage}
	&
		\begin{minipage}{.5\textwidth}
			\normalsize
			\begin{center}
\href{https://orcid.org/0000-0003-4974-4134}{\includegraphics[scale=0.06]{orcid.pdf}\hspace{1mm}\textbf{Arne Sachtler}} \\
	School of Computation, Information and Technology\\
	Technical University of Munich (TUM)\\
	Garching, Germany\\
	\texttt{arne.sachtler@dlr.de}\\
			\end{center}
		\end{minipage}
	\end{tabular}\hspace{100cm}
\thanks{\textcolor{blue}{
   This is a preprint of the following chapter: Alin Albu-Schäffer and Arne Sachtler, \emph{What Can Algebraic Topology and Differential Geometry Teach Us About Intrinsic Dynamics and Global Behavior of Robots?}, published in \emph{Robotics Research}, edited by {Aude Billard, Tamim Asfour, and Oussama Khatib}, \textbf{2023}, Springer reproduced with permission of Springer Nature Switzerland AG. The final authenticated version is available online at: \url{https://doi.org/10.1007/978-3-031-25555-7_32}.
}}
}

\title{What Can Algebraic Topology and Differential Geometry Teach Us About Intrinsic Dynamics and Global Behavior of Robots?}

\begin{document}
\maketitle

\begin{abstract}
Traditionally, robots are regarded as universal motion generation machines.
They are designed mainly by kinematics considerations while the desired dynamics is imposed by strong actuators and high-rate control loops.
As an alternative, one can first consider the robot's intrinsic dynamics and optimize it in accordance with the desired tasks.
Therefore, one needs to better understand intrinsic, uncontrolled dynamics of robotic systems.
In this paper we focus on periodic orbits, as fundamental dynamic properties with many practical applications.
Algebraic topology and differential geometry provide some fundamental statements about existence of periodic orbits.
As an example, we present periodic orbits of the simplest multi-body system: the double-pendulum in gravity.
This simple system already displays a rich variety of periodic orbits.
We classify these into three classes: toroidal orbits, disk orbits and nonlinear normal modes.
Some of these we found by geometrical insights and some by numerical simulation and sampling.
\end{abstract}

\section{Introduction}
The traditional approach to robot motion generation is to first plan trajectories on a kinematic level and then develop controllers for tracking the planned trajectories.
The robot hardware, and therefore its dynamics, are considered to be given a priori. 
As the robot is understood as a universal motion generation machine, the ideal controller must track any trajectory to the best extent possible, leading to the ideal of a fully decoupling controller.
This compensates the intrinsic dynamics and leads to simple error dynamics; for example, fully decoupled, second-order linear differential equations in computed torque and operational space control~\cite{Khatib1987,DeLuca1998}.
Theoretically, one could control an elephant to jump like a flea this way. 
Despite the huge success of this approach in robotics, the limitations are also obvious and well known: actuator saturation, model errors, and unmodelled dynamics lead, in the extreme case, to severe performance limitations, robustness problems up to instability, and high energy consumption.

An alternative perspective has been taken in robotics for a long time as well, leading to minimalistic and passivity-based control~\cite{Hogan1985,AlbuSchaeffer2007a}, bio-inspired design, embodied intelligence, passive walkers~\cite{McGeer1990,Weele2001}, and locomotion template anchoring~\cite{Holmes2006}.
When operating a robot on trajectories matching its natural dynamics, one only needs very small control action. 
Such trajectories correspond to geodesics with respect to the Jacobi metric. 
Also research in neuroscience suggests that humans operate their arms on geodesics~\cite{Biess2007,Hogan2012}.
Especially, for robots that are supposed to perform quite specific, periodic, or quasi-periodic motions most of the time, such as in legged locomotion, linking the design of the robot and its intrinsic dynamics to its desired task promises benefits in terms of performance and energy efficiency. 
Paraphrasing Rodney Brooks~\cite{Brooks1990}, we would say \emph{Elephants don't play hopscotch either}. 
Also industrial robots, when used in large production lines, perform mostly very repetitive motions. 
Energy efficiency becomes relevant here as well, in the context of $CO_2$ neutrality, while maximizing speed and/or force is always the central concern. 

In this paper we contribute some insights into understanding intrinsic robot dynamics as methodologies to generate highly efficient motions. 
To this end, we go back to the roots of mechanics, taking a closer look at the principles of least action and interpreting them geometrically.
Although these principles are dating back to Maupertuis, Euler, Lagrange, Jacobi, and Hamilton, almost three centuries of developments in differential geometry, algebraic topology, and of numerical methods, make it worth taking a fresh look at their meaning and implications. 
We address motions that can be performed at constant total energy, in particular periodic motions. 
Although our intuition of frictionless, constant energy behavior of robot dynamics is that of chaotic or at least very complicated motions, it turns out that there are even more periodic, regular motions than in general linear systems. 
As an example, we will present the zoo of intrinsic periodic motions of the double pendulum, i.e., of the most basic, 2 DoF robot dynamics.

\subsection{A Very Short Primer on Robot Dynamics}
The classical way of deriving the equations of motion of mechanical systems is based on the Lagrange formalism~\cite{Lynch2017,Murray1994,Arnold1989,Sussman2001}.
One defines a Lagrangian 
\begin{equation}\label{eq:lagrangian}
  L(\q, \dqq) = T(\q, \dqq) - U(\q)
\end{equation}
as the difference of kinetic energy $T(\q, \dqq)$ and potential energy $U(\q)$, with $\q \in \mathcal{Q}$ being configuration variables and $\mathcal{Q}$ the configuration space.
We integrate the Lagrangian over \emph{candidate} trajectories $\qq(t)$ using the Hamiltonian action integral 
\begin{equation}\label{eq:action}
  S_H(\q) = \int_{t_1}^{t_2} L(\q, \dqq)\d t.
\end{equation}
The action integral is a functional: it takes an input function of a single variable and outputs a real number.
For the Hamiltonian action integral, this input is a function $\q: \mathbb{R} \rightarrow \mathcal{Q}$ of time and the output is the total action of the mechanical system on $\q(t)$ from $t_1$ to $t_2$. 
Then we take Hamilton's Principle of Least Action to select a true system trajectory $\hat{\qq}(t)$ out of the canditates:
\begin{quote}
The system takes a trajectory $\hat{\q}(t)$ between $\q_1=\q(t_1)$ and $\q_2=\q(t_2)$, that extremizes $S_H(\hat{\q})$ over all possible smooth paths satisfying the given boundary conditions.
\end{quote}
Using variational calculus, the extremizer for (\ref{eq:action}) locally satisfies the Euler-Lagrange equations
\begin{equation}\label{eq:euler_lagrange_eq}
  \frac{\d}{\d t}\left(\frac{\partial L}{\partial \dqq}\right) - \frac{\partial L}{\partial \q} = \nnull.
\end{equation}
The resulting equations of motion can be, in absence of external generalized forces, written in the well-known standard form
\begin{equation}\label{eq:mcg}
  \MM(\qq)\ddqq + \CC(\q, \dqq)\dqq + \gggg(\qq) = \nnull
\end{equation}
with mass matrix $\MM(\qq)$, potential forces $\gggg(\qq)$ and Coriolis and centrifugal forces $\CC(\q, \dqq)\dqq$.
From this step on, one classically only uses (\ref{eq:mcg}) for analyzing the dynamics of the multi-body system.
The power and large variety of applications make us not spend too many thoughts on the meaning of the initial action integral (\ref{eq:action}), which is merely considered a creative intermediate step needed to arrive at the Euler-Lagrange equations.

We would like to go one step back and introduce some classical results directly derived from an alternative version of the principle of least action: we look at the Maupertuis-Euler-Lagrange-Jacobi formulation. 
We will take advantage of this in gaining new insights into the intrinsic dynamics of conservative mechanical systems, especially regarding the existence and numerical computation of periodic trajectories of various types. 
Although this body of work, leading to some remarkable insights, is still today a topic of active research in mathematical physics and theoretical mechanics, it seems to be largely unknown to the robotics community. 
We believe that the theoretical results of the last decades as well as the powerful numerical tools and the computation power available today might lead to many applications in robotics.

\subsection{Mauptertuis' Principle of Least Action}
If the Hamiltonian $H(\q, \dqq)$, which in robotics is identical to the total energy, stays constant during motion, time can be completely eliminated from Hamilton's principle of least action, leading to Maupertuis' principle
\begin{equation}\label{eq:Maupertuis_action}
  S_M(\q) = \int_{q_1}^{q_2} \pp \d \q,
\end{equation}
where $\pp= \nicefrac{\partial L}{\partial\dqq}$ is the generalized momentum, expressed as a function of position along the trajectory of constant energy~\cite{Arnold1989}. 
These two principles of least action can be derived from each other in the case of constant energy~\cite{Gray1996}.
The elimination of time, and thus of velocities, has two major implications:
\begin{enumerate}
    \item the search for trajectories of the systems can be performed in the $n$-dim\-en\-sio\-nal configuration space instead of the $2n$-dimensional phase space; and
    \item one can access a huge body of results from Riemannian geometry and algebraic topology.  
\end{enumerate}
If parametrizing the curve by time, \eqref{eq:Maupertuis_action} will take the form
\begin{equation}\label{eq:Maupertuis_action_time}
  S_M = \int_{t_1}^{t_2} 2T(\qq,\dqq) \d t.
\end{equation}
Let's compare: in Hamilton's principle of least action we fix the endpoints $\qq_{1}$, $\qq_{2}$ \emph{and} the corresponding times $t_1$, $t_2$, but we do not fix the total energy. We find \emph{trajectories} of the system using this principle.
In Maupertuis' principle we fix the endpoints $\qq_{1}$, $\qq_{2}$ and the total energy, but do not care about times. We find \emph{configuration paths} only, without velocity information. We can, however, reconstruct time and velocity from the configuration path considering the fixed total energy.
For a purely geometric formulation of conservative motions, geodesics play a central role.
We introduce them now.

\subsection{Geodesics on Riemannian Manifolds}
\begin{figure}
  \centering
  \def\svgwidth{.3\textwidth}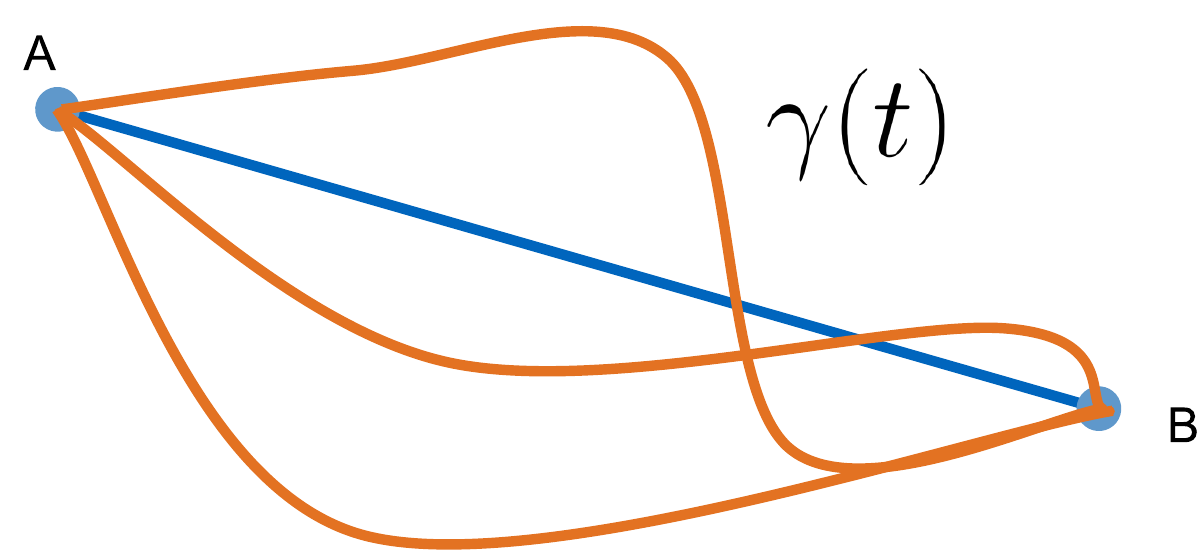
  \caption{Geodesic on Euclidean space. Blue shows a geodesic and orange non-geodesics.}\label{fig:geodesics}
\end{figure}

The notion of geodesics is one of the most basic concepts in differential geometry~\cite{Needham2021}.
Let $(\mathcal{M},g)$ be a Riemannian manifold\footnote{We cannot introduce manifolds here. Great starting points are the books by T. Needham on differential geometry~\cite{Needham2021}, and by F. Morgan on Riemannian geometry~\cite{Morgan2020}.} with metric $g_{ij}$.
A geodesic is the \emph{straightest} curve between two points \cite{Needham2021}.
Let $\gamma: [s_1, s_2] \rightarrow \mathcal{M}$ be a parametric curve.
It will be called geodesic if it extremizes the arc-length integral\footnote{We use Einstein notation in this paper: whenever one up-down pair of indices match, we implicitly sum over them. Example: $g_{ij}a^i b^j := \sum_i \sum_j g_{ij}a^i b^j$}
\begin{equation}\label{eq:arc_length}
  L(\gamma) = \int_{s_1}^{s_2}\sqrt{g_{ij}{\gamma'}^i {\gamma'}^j}\d s,
\end{equation}
with $ \gamma'  = \nicefrac{\d \gamma}{\d s}$.
In general the extremum does not need to be a minimum. 
For example, on a sphere, geodesics are both segments of the great circle passing through two given points, which is unique if the points are not antipodal. 
Fig.~\ref{fig:geodesics} illustrates the principle in the Euclidean space $(\mathbb{R}^{2}, \delta_{ij})$.
For this example, the blue straight line is the globally shortest path and the only geodesic.
Geodesics, and thus extremizers of (\ref{eq:arc_length}), satisfy the \emph{geodesic equation}~\cite{Sussman2012,Lee2018}
\begin{equation}\label{eq:geodesic_eq}
  \frac{\partial^2 \gamma^a}{\partial s^2} + \mathrm{\Gamma}^{a}_{bc} \frac{\partial \gamma^b}{\partial s}\frac{\partial \gamma^c}{\partial s} = 0,
\end{equation}
where $\mathrm{\Gamma}^{a}_{bc}$ are Christoffel symbols of the second kind derived solely from the metric \begin{equation}\label{eq:christoffel_2nd}
  \mathrm{\Gamma}^{a}_{bc} = \frac{1}{2}g^{ai}\left(\frac{\partial }{\partial x^{c}}g_{ib} + \frac{\partial}{\partial x^{b}}g_{ic} - \frac{\partial}{\partial x^i}g_{bc}\right),
\end{equation}
where $g^{ai}$ is the inverse of $g_{ai}$, i.e., $g^{i\alpha}g_{\alpha j}={\delta^{i}}_j$.

\subsection{A Purely Geometric Perspective on Lagrangian Mechanics}
Starting from Maupertuis' principle of least action \eqref{eq:Maupertuis_action}, \eqref{eq:Maupertuis_action_time}, it can be shown that trajectories of constant energy between two points are geodesics with respect to the Jacobi metric~\cite{Arnold1989,DiCairano2019,Casetti2000} 
\begin{equation}\label{eq:jacobi_metric}
 {{}^J g}_{ij} = 2(E - U(\qq))m_{ij},
\end{equation}
where $m_{ij}$ is the inertia tensor.
The derivation of the Jacobi metric is based on this insight: if energy is constant ($T(\qq, \dqq) + U(\qq) = E$), then velocity can be expressed as a function of position on any trajectory.
So let us parametrize a motion along the curve $\gamma$ instead of time $t$ by arc length $s$, which is in bijective relation $s(t)$ to time. With the notation $\q'=\nicefrac{\d \qq}{\d s}$ we have:
\begin{equation}
    2T(\qq, \dqq)=m_{ij}\dq^i\dq^j=m_{ij} q'^i q'^j \left(\frac{\d s}{\d t}\right)^2 =2(E-U(\qq)).
\end{equation}
This relates the differentials $\d t$ and $\d s$
\begin{equation}\label{eq_Jacobi_metric_reparametrization}
    \d t = \sqrt{\frac{m_{ij} q'^i q'^j }{2(E-U(\qq))}} \d s.
\end{equation}
Using $2T = 2(E-U)$ allows to remove $T$ from \eqref{eq:Maupertuis_action_time}.
We also substitute $t$ with the curve parameter $s$ and get
\begin{eqnarray}
   S_M&= & \int_{t_1}^{t_2} 2T(\qq, \dqq) \d t \nonumber \\ &=& \int_{s_1}^{s_2} 2(E-U(\qq)) \sqrt{\frac {m_{ij} q'^i q'^j} {2(E-U(\qq))}} \d s \nonumber\\
    & = & \int_{s_1}^{s_2} \sqrt{2(E-U(\qq))m_{ij} q'^i q'^j} \d s \nonumber \\ &=&
    \int_{s_1}^{s_2} \sqrt{{{}^J g}_{ij} q'^i q'^j} \d s,
\end{eqnarray}
which is indeed exactly the arc length with respect to the Jacobi metric \eqref{eq:jacobi_metric}.
This is remarkable!
Isoenergetic trajectories of the multi-body system in a potential field are geodesics on the manifold $(\mathcal{Q}, {{}^J g})$, where $\mathcal{Q}$ is the configuration space.
This eliminates velocities from the problem - it is purely about curves on the configuration space.
The solutions describe only the path itself, not the timing along it. 
Velocities are obtained by scaling the tangent according to $\dqq = \qq' \nicefrac{\d s}{\d t}$: we need to scale the tangent to satisfy the constant energy condition.
We can also reconstruct time by integrating~\eqref{eq_Jacobi_metric_reparametrization}.

For potential-free systems the metric is proportional to the inertia tensor ${{}^J g}_{ij}= 2E m_{ij}$ and the constant factor can be ignored when searching for geo\-desics. Thus, for potential-free rigid body systems, trajectories are geodesics w.r.t.\ the inertia tensor.
In this case, the paths are independent of the energy, varying the energy only changes the speed used to trace out the geodesics in configuration space.
In contrast, for systems in potential fields, the geodesics generally vary with the total energy $E$, as the Jacobi metric is energy-dependent. 
The necessary and sufficient conditions for the paths to be independent of energy also in presence of potential energy have been derived in~\cite{AlbuSchaeffer2021}.

\subsection{Algorithms: Shooting versus Extremizing Arc Length}
How can we use the insights presented so far to find intrinsic paths of dynamic systems? The basic example is still the path of constant energy between two points (Fig.~\ref{fig:geodesics}), which can be, however, easily extended to periodic paths of line topology and to closed paths. The approach mostly used in robotics for finding a path between the points $\qq_A$ and $\qq_B$ in Fig.~\ref{fig:geodesics} is to choose an initial guess for the velocity $\dqq_A$ and "shoot" from the initial state $(\qq_A, \dqq_A)$, i.e. simulate the robot dynamics under additional constraints of time, energy, etc. A measure of the amount by which the point $\qq_B$ is missed by the path is fed to the optimization algorithm, which will adapt the initial velocity $\dqq_A$ until it will hopefully hit the point $\qq_B$. Obviously, for long paths, the problem is not very well conditioned and there exist many improvements, for example by multiple shooting algorithms. For periodic trajectories, the Poincaré map gives a similar procedure for optimization, based on simulating the system dynamics. As a basic principle, one optimizes in the space of curves which are feasible solutions of the differential equation, trying to satisfy the boundary conditions.

In contrast, by using the principle of least action, one optimizes in the space of curves satisfying the boundary conditions, i.e. passing through the points $\qq_A$ and $\qq_B$,  but which are not necessarily solutions of the system's differential equations (yet). The algorithms then make these curves system trajectories by zeroing the amount by which the curves fail to satisfy the geodesic equation. Equivalently, the algorithm extremizes the arc length in the corresponding Jacobi metric, which can be intuitively thought of as contracting rubber strings on the manifold. We look in more detail at this algorithm in the following.

\section{String Relaxation}\label{sec:curve shortening}
Imagine you take a string and fix the two ends to two distinct points.
Now think of the string as a rubber band: it will naturally contract to the (locally) shortest possible path between the two endpoints.

Suppose we have a not (yet) geodesic curve $\gamma(s, t)$, which we would like to converge to a geodesic as $t \rightarrow \infty$.
We take the geodesic equation and make $\gamma(s, t)$ satisfy it more and more over time by the PDE 
\begin{equation}\label{eq:string_pde}
  \frac{\partial \gamma^a}{\partial t} = \frac{\partial^2 \gamma^a}{\partial s^2} + \mathrm{\Gamma}^a_{bc} \frac{\partial \gamma^b}{\partial s}\frac{\partial \gamma^c}{\partial s}.
\end{equation}

\begin{figure}
  \centering
  \def\svgwidth{.5\textwidth}
  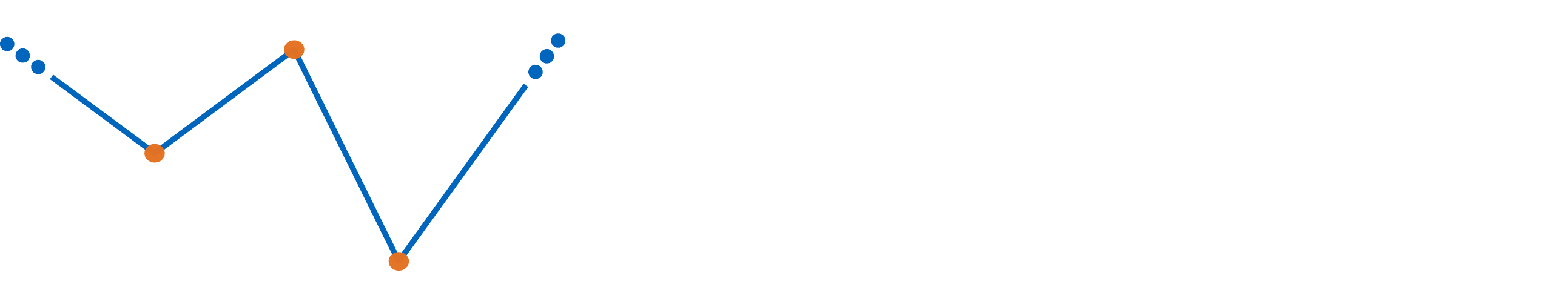
  \caption{Discrete string in Euclidean space}
  \label{fig:strings}
\end{figure}

Let's discretize $\gamma(s, t)$ in space and time and write it as $\gamma_t(k) = \gamma(k\Delta s, t\Delta T)$ (Fig.~\ref{fig:strings}).
We use central differences for the first derivatives and also discretize the second derivative.
This results in the update rule \begin{align}\label{eq:update_rule}
  \gamma_{t+1}^a(k) &= \gamma_t^a(k) + \Delta T \frac{\gamma_t^a(k+1) - 2\gamma_t^a(k) + \gamma_t^a(k-1)}{(\Delta s)^2} \nonumber \\ 
  &+ \Delta T \, \mathrm{\Gamma}^a_{bc}\frac{\gamma_t^b(k+1)-\gamma_t^b(k-1)}{2\Delta s}\frac{\gamma_t^c(k+1) - \gamma_t^c(k-1)}{2\Delta s}.\nonumber
\end{align}

Note that we show here an explicit Euler scheme  for discretizing (\ref{eq:string_pde}).
This is not what one would implement in practice, but serves to illustrate the idea.
Convergence is rather slow for the explicit scheme as small time steps must be chosen.
For this kind of relaxation dynamics much faster convergence can be obtained by switching to implicit solvers allowing to crank up $\Delta T$ a lot~\cite{Grinspun2008a}.

\subsection{Example: Dynamical System}
\begin{figure*}
  \centering
  \subfloat[Evolution of Strings]{\def\svgwidth{.3\textwidth}\tiny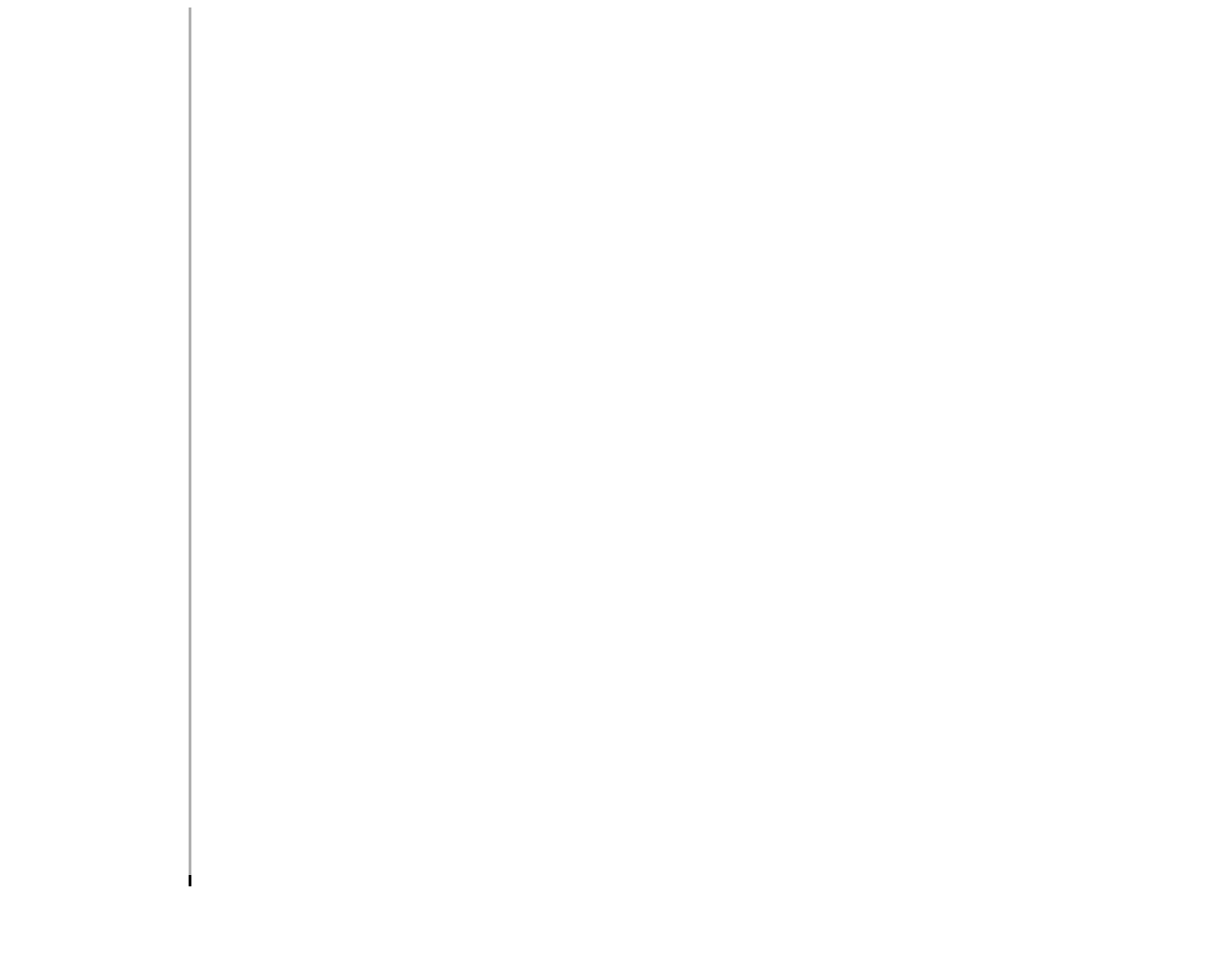}\hfill
  \subfloat[Length\label{fig:traj_res:length}]{\def\svgwidth{.32\textwidth}\tiny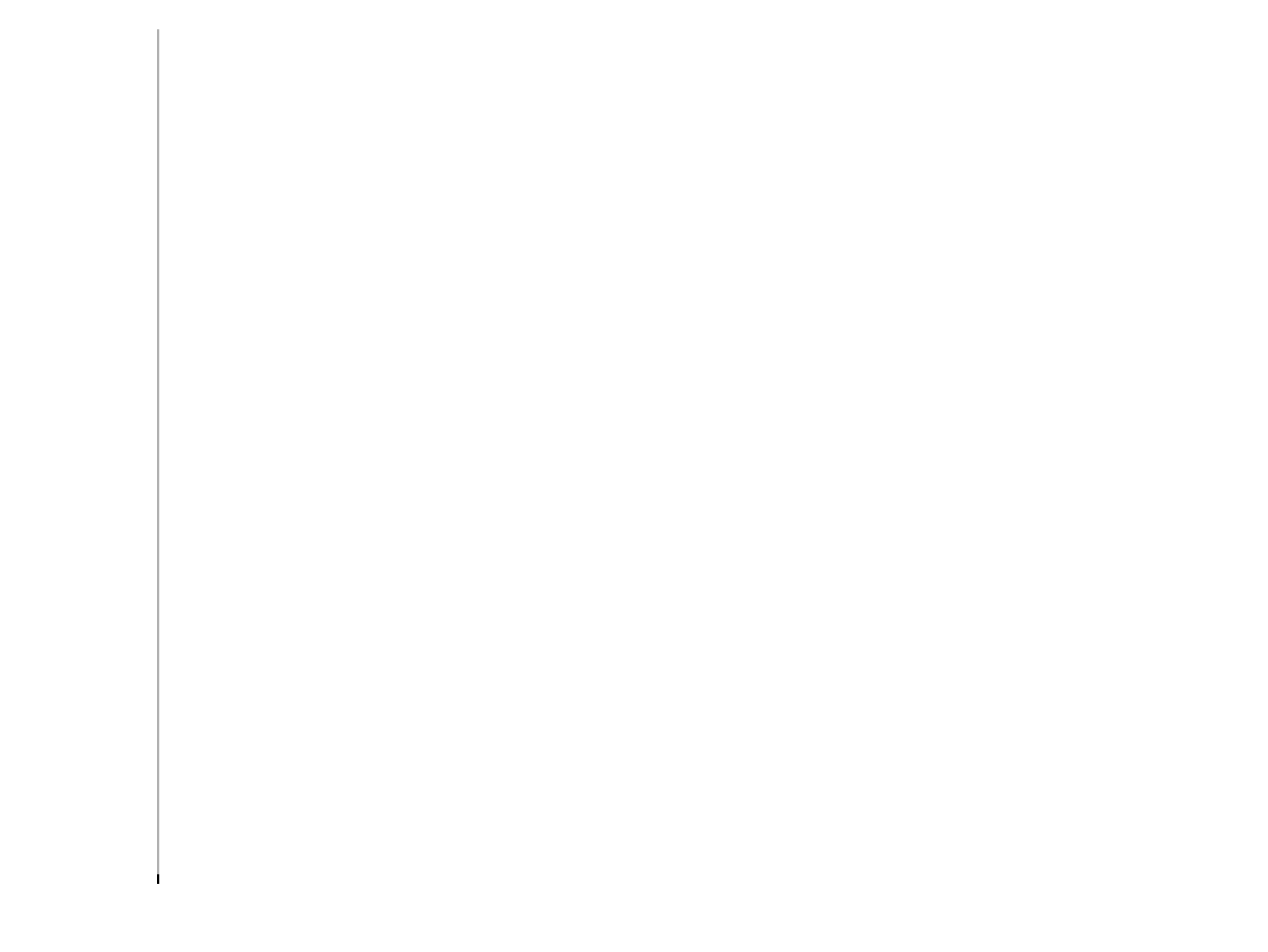}\hfill
  \subfloat[Simulation\label{fig:traj_res:sim}]{\def\svgwidth{.32\textwidth}\tiny%% Creator: Inkscape 1.2.2 (b0a8486541, 2022-12-01), www.inkscape.org
%% Accompanies image file 'traj_phase.pdf' (pdf, eps, ps)
\begingroup%
  \makeatletter%
  \providecommand\color[2][]{%
    \errmessage{(Inkscape) Color is used for the text in Inkscape, but the package 'color.sty' is not loaded}%
    \renewcommand\color[2][]{}%
  }%
  \providecommand\transparent[1]{%
    \errmessage{(Inkscape) Transparency is used (non-zero) for the text in Inkscape, but the package 'transparent.sty' is not loaded}%
    \renewcommand\transparent[1]{}%
  }%
  \providecommand\rotatebox[2]{#2}%
  \newcommand*\fsize{\dimexpr\f@size pt\relax}%
  \newcommand*\lineheight[1]{\fontsize{\fsize}{#1\fsize}\selectfont}%
  \ifx\svgwidth\undefined%
    \setlength{\unitlength}{439.28186035bp}%
    \ifx\svgscale\undefined%
      \relax%
    \else%
      \setlength{\unitlength}{\unitlength * \real{\svgscale}}%
    \fi%
  \else%
    \setlength{\unitlength}{\svgwidth}%
  \fi%
  \global\let\svgwidth\undefined%
  \global\let\svgscale\undefined%
  \makeatother%
  \begin{picture}(1,0.73664834)%
    \lineheight{1}%
    \setlength\tabcolsep{0pt}%
    \put(0,0){\includegraphics[width=\unitlength,page=1]{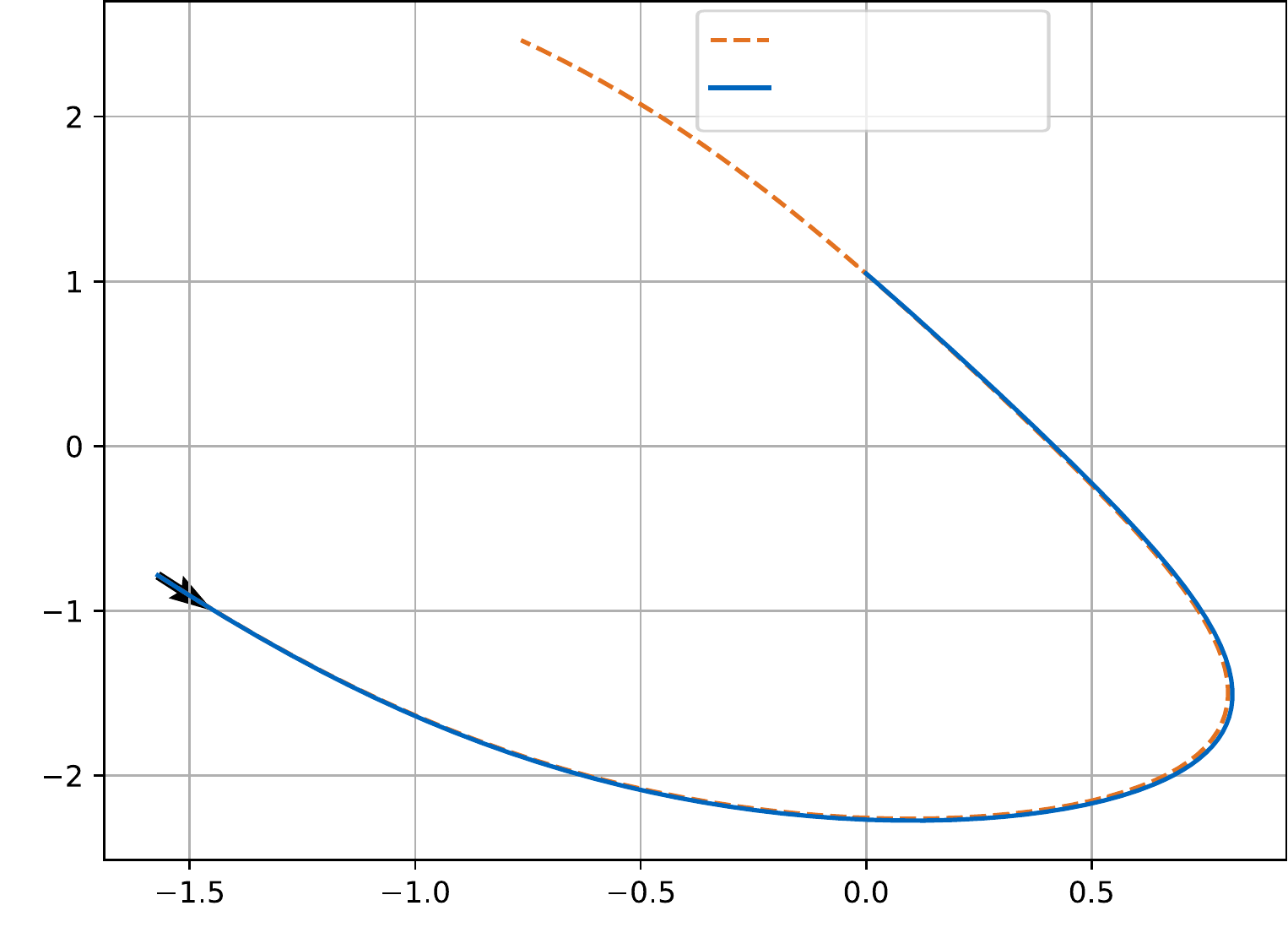}}%
    \put(0.61352077,0.69739213){\color[rgb]{0,0,0}\makebox(0,0)[lt]{\lineheight{1.25}\smash{\begin{tabular}[t]{l}Simulation\end{tabular}}}}%
    \put(0.61118048,0.66281538){\color[rgb]{0,0,0}\makebox(0,0)[lt]{\lineheight{1.25}\smash{\begin{tabular}[t]{l}Geodesic\end{tabular}}}}%
    \put(0.48082854,0.00663807){\color[rgb]{0,0,0}\makebox(0,0)[lt]{\lineheight{1.25}\smash{\begin{tabular}[t]{l}$q_1$ [rad]\end{tabular}}}}%
    \put(0.01805164,0.34683583){\color[rgb]{0,0,0}\rotatebox{90}{\makebox(0,0)[lt]{\lineheight{1.25}\smash{\begin{tabular}[t]{l}$q_2$ [rad]\end{tabular}}}}}%
    \put(0,0){\includegraphics[width=\unitlength,page=2]{traj_phase.pdf}}%
    \put(0.69156223,0.54547415){\color[rgb]{0,0,0}\makebox(0,0)[lt]{\lineheight{1.25}\smash{\begin{tabular}[t]{l}$\q_B$\end{tabular}}}}%
    \put(0.09258437,0.31809261){\color[rgb]{0,0,0}\makebox(0,0)[lt]{\lineheight{1.25}\smash{\begin{tabular}[t]{l}$\q_A$\end{tabular}}}}%
  \end{picture}%
\endgroup%
}
  \caption{String relaxation to find trajectories of dynamical systems}
  \label{fig:traj_results}
\end{figure*}

Let us next try string relaxation on a dynamical system.
We consider the configuration space of a double pendulum and choose a total energy $E$.
Assume we want a trajectory from a configuration $\q_A$ to $\q_B$.
For constant energy, we can fully capture its dynamics in the Jacobi metric and make the problem purely geometric.
We fix a simulated string at the two configurations $\q_A$ and $\q_B$ and let it contract under the Jacobi metric.
Fig.~\ref{fig:traj_results} shows one such example.
The dashed blue line in Fig.~\ref{fig:traj_results} shows the initial string.
Over the iterations the string converges to the orange curve.
At the same time, the Riemannian length of the string decreases (blue in Fig.~\ref{fig:traj_res:length}).
In orange we show the convergence velocity of the relaxation measured by $v(t) = \sum_k ||\gamma_{t-1}(k) - \gamma_{t}(k)||$.

Once the string has converged to a geodesic $\ggamma(s)$ we compare it to a forward simulated system.
One last step is to determine the initial velocity $\qdot_A$.
We scale the tangent of $\ggamma(s)$ to match the physical energy using the inverse of (\ref{eq_Jacobi_metric_reparametrization}).
We scale the tangent of $\ggamma(s)$ such that the total energy is preserved \begin{equation}
  \qdot_A = \sqrt{\frac{2(E - U(\q_A))}{m_{ij}{\gamma'}_A^i {\gamma'}_A^j}} \frac{\partial \ggamma}{\partial s}.
\end{equation}
Starting from the state $(\q_A, \dqq_A)$ we simulate the double pendulum using a Runge-Kutta integration scheme for some time and obtain the curve shown in orange in Fig.~\ref{fig:traj_res:sim}.
We observe that the simulated trajectory follows the relaxed string in configuration space.
As we still have energy at $\q_B$ we pass by it and continue.

In the example we have fixed the string at two fixed end-points $\qq_A$ and $\qq_B$.
This allows us to find geodesics connecting two configurations.
But nobody stops us from closing the string by connecting the first and the last vertex on the string.
Then, the string has no boundary conditions that would hold it in place.
Often it will collapse to a point, but sometimes the topology of the space prevents this collapse - we will look into this phenomenon in the next section.
In that case we can use the string relaxation to find periodic orbits.

\section{Topological Insights into Periodic Orbits}\label{sec:algebraic-topology}
\begin{figure}
  \centering
  \includegraphics[width=.5\textwidth]{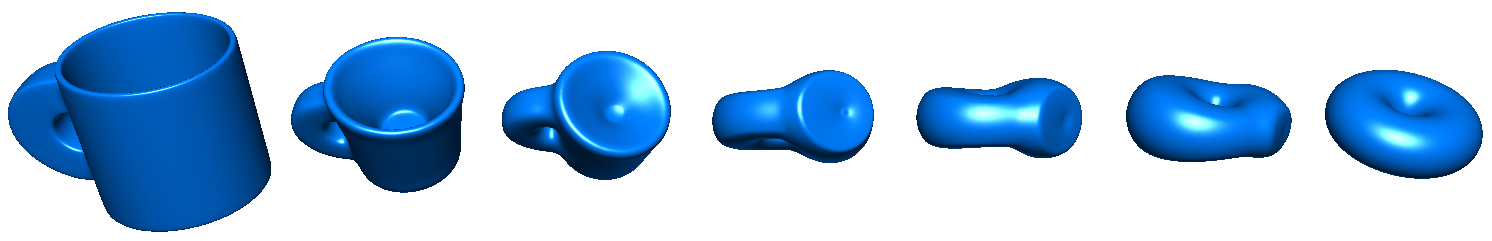} 
  \caption{The mug can be continuously deformed into a doughnut; they have the same topology. Image rendered from the 3D model provided by \cite{Segerman2016}.}
  \label{fig:homotopy-torus-mug}
\end{figure}

Besides providing algorithms to find intrinsic trajectories, the principle of least action leads to the possibility of making general theoretical predictions about the types and numbers of special trajectories, such as periodic orbits. To get a flavor of the approaches, one needs some topology, in particular algebraic topology. 
We will only show a brief summary of the most essential concepts here. 
The main interest of algebraic topology is to classify manifolds into certain equivalence classes and find invariant quantities which uniquely characterize them \cite{Cartan1956}. 
Manifolds are homotopy-equivalent if they can be smoothly deformed into each other by a homotopy \cite{Hatcher2002,Lee2011}. In Fig.~\ref{fig:homotopy-torus-mug}, the doughnut and the mug are homotopy-equivalent and this is, intuitively speaking because they are both 2-dimensional, closed, unbounded surfaces with one hole.

An effective way to classify $n$-dimensional manifolds is to count their number of holes of dimensions $0, ..., n$, which are described by the Betti numbers $b_0,...,b_n$. For 2-dimensional surfaces, a zero-dimensional hole is a gap between two path-connected components. So, for example, $b_0(\mathbb{S}^2)=1$ for a sphere $\mathbb{S}^2$ and $b_0=k$ for a manifold composed of $k$ disjoint spheres.

\begin{figure}
  \centering
  \def\svgwidth{0.15\textwidth}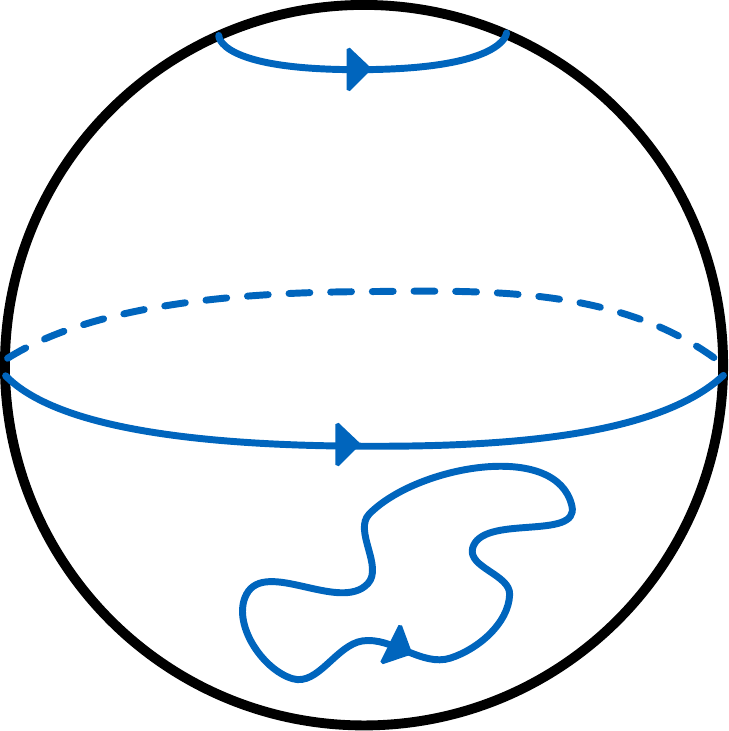\hspace{1cm}
  \def\svgwidth{0.25\textwidth}%% Creator: Inkscape 1.2.2 (b0a8486541, 2022-12-01), www.inkscape.org
%% Accompanies image file 'closed_curves_torus.pdf' (pdf, eps, ps)
\begingroup%
  \makeatletter%
  \providecommand\color[2][]{%
    \errmessage{(Inkscape) Color is used for the text in Inkscape, but the package 'color.sty' is not loaded}%
    \renewcommand\color[2][]{}%
  }%
  \providecommand\transparent[1]{%
    \errmessage{(Inkscape) Transparency is used (non-zero) for the text in Inkscape, but the package 'transparent.sty' is not loaded}%
    \renewcommand\transparent[1]{}%
  }%
  \providecommand\rotatebox[2]{#2}%
  \newcommand*\fsize{\dimexpr\f@size pt\relax}%
  \newcommand*\lineheight[1]{\fontsize{\fsize}{#1\fsize}\selectfont}%
  \ifx\svgwidth\undefined%
    \setlength{\unitlength}{380.25000865bp}%
    \ifx\svgscale\undefined%
      \relax%
    \else%
      \setlength{\unitlength}{\unitlength * \real{\svgscale}}%
    \fi%
  \else%
    \setlength{\unitlength}{\svgwidth}%
  \fi%
  \global\let\svgwidth\undefined%
  \global\let\svgscale\undefined%
  \makeatother%
  \begin{picture}(1,0.57790927)%
    \lineheight{1}%
    \setlength\tabcolsep{0pt}%
    \put(0,0){\includegraphics[width=\unitlength,page=1]{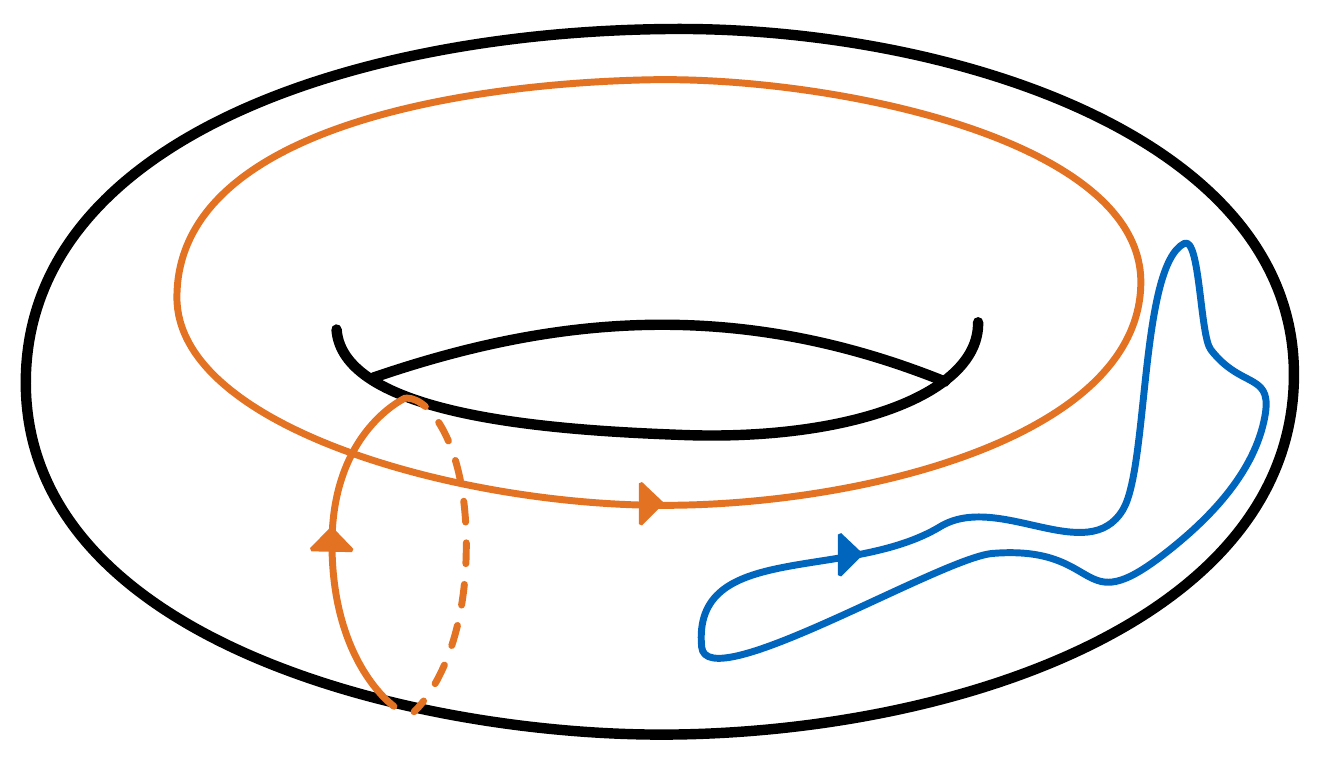}}%
    \put(0.27459667,0.14666736){\color[rgb]{0,0,0}\makebox(0,0)[lt]{\lineheight{1.25}\smash{\begin{tabular}[t]{l}$a$\end{tabular}}}}%
    \put(0.41371826,0.13103374){\color[rgb]{0,0,0}\makebox(0,0)[lt]{\lineheight{1.25}\smash{\begin{tabular}[t]{l}$b$\end{tabular}}}}%
  \end{picture}%
\endgroup%

  \caption{Classes of closed curves: cycles on the sphere and the torus. Blue curves can be continuously collapsed to a point; orange curves cannot.}
  \label{fig:cycles_on_sphere&Torus}
\end{figure}

One-dimensional holes are found and counted by classes of closed curves (1-cycles) on the manifold, which cannot be shrunk to a point. 
For example, on a sphere $\mathbb{S}^2$ all closed curves can be shrunk to a point (Fig.~\ref{fig:cycles_on_sphere&Torus}), so $b_1(\mathbb{S}^2)=0$. 
On a torus $\mathbb{T}^2$, there are two distinct classes of curves that cannot be shrunk to a point, as shown in Fig.~\ref{fig:cycles_on_sphere&Torus} in orange. Therefore, on the torus $b_1(\mathbb{T}^2)=2$. In each of the two classes, there are infinitely many curves, which can be continuously deformed into each other. However, note that the curves from class $a$ cannot be continuously deformed into curves of class $b$, the two sets are disjoint.
Finally, both the sphere and the torus have one 2-dimensional hole (whose boundary is a closed surface), and therefore, $b_2(\mathbb{S}^2)=b_2(\mathbb{T}^2)=1$.

For the analysis of periodic, closed paths of a robot, the first Betti number, related to families of closed curves, is of particular interest. 
Consider a double pendulum (2 DoF vertical manipulator with gravity).
Its configuration space is the 2-torus $\mathbb{T}^2$.
The two distinct types of cycles $a$ and $b$ form a basis (independent generators) for the group of all possible cycles. 
The neutral element of the group is the zero cycle, i.e., the set of all curves which can be shrunk to a point. The composition $a + b$ of two elements of the group means that the cycles are just followed one after the other, and, e.g., $2a + 3b$ means that a curve winds two times around the first joint and three times around the second joint. 
The order does not matter, the group is considered abelian (commutative). 
This group is called the homology group ${\cal{H}}_1(\mathbb{T}^2)$ and its rank is indicated by the Betti number $b_1$. 
Any closed curve on the torus can be generated starting from $\alpha a + \beta b$ with $\alpha, \beta \in \mathbb{Z}$ and through a homotopy (continuous deformation). Note, again, that for different $\alpha, \beta$, the curves cannot be continuously deformed into each other, the classes are disjoint. 

This will be used in Section~\ref{sec:m-n-orbits} to directly show that there are infinitely many periodic closed orbits for the double pendulum, at least one for each element of the homology group  ${\cal{H}}_1(\mathbb{T}^2)$.

\section{A Case Study: Periodic Orbits of the Double Pendulum}
\begin{figure}
  \centering
  \subfloat[Double Pendulum\label{fig:the_system}]{\def\svgwidth{.15\textwidth}%% Creator: Inkscape 1.2.2 (b0a8486541, 2022-12-01), www.inkscape.org
%% Accompanies image file 'dp_sym.pdf' (pdf, eps, ps)
\begingroup%
  \makeatletter%
  \providecommand\color[2][]{%
    \errmessage{(Inkscape) Color is used for the text in Inkscape, but the package 'color.sty' is not loaded}%
    \renewcommand\color[2][]{}%
  }%
  \providecommand\transparent[1]{%
    \errmessage{(Inkscape) Transparency is used (non-zero) for the text in Inkscape, but the package 'transparent.sty' is not loaded}%
    \renewcommand\transparent[1]{}%
  }%
  \providecommand\rotatebox[2]{#2}%
  \newcommand*\fsize{\dimexpr\f@size pt\relax}%
  \newcommand*\lineheight[1]{\fontsize{\fsize}{#1\fsize}\selectfont}%
  \ifx\svgwidth\undefined%
    \setlength{\unitlength}{566.92913386bp}%
    \ifx\svgscale\undefined%
      \relax%
    \else%
      \setlength{\unitlength}{\unitlength * \real{\svgscale}}%
    \fi%
  \else%
    \setlength{\unitlength}{\svgwidth}%
  \fi%
  \global\let\svgwidth\undefined%
  \global\let\svgscale\undefined%
  \makeatother%
  \begin{picture}(1,1)%
    \lineheight{1}%
    \setlength\tabcolsep{0pt}%
    \put(0,0){\includegraphics[width=\unitlength,page=1]{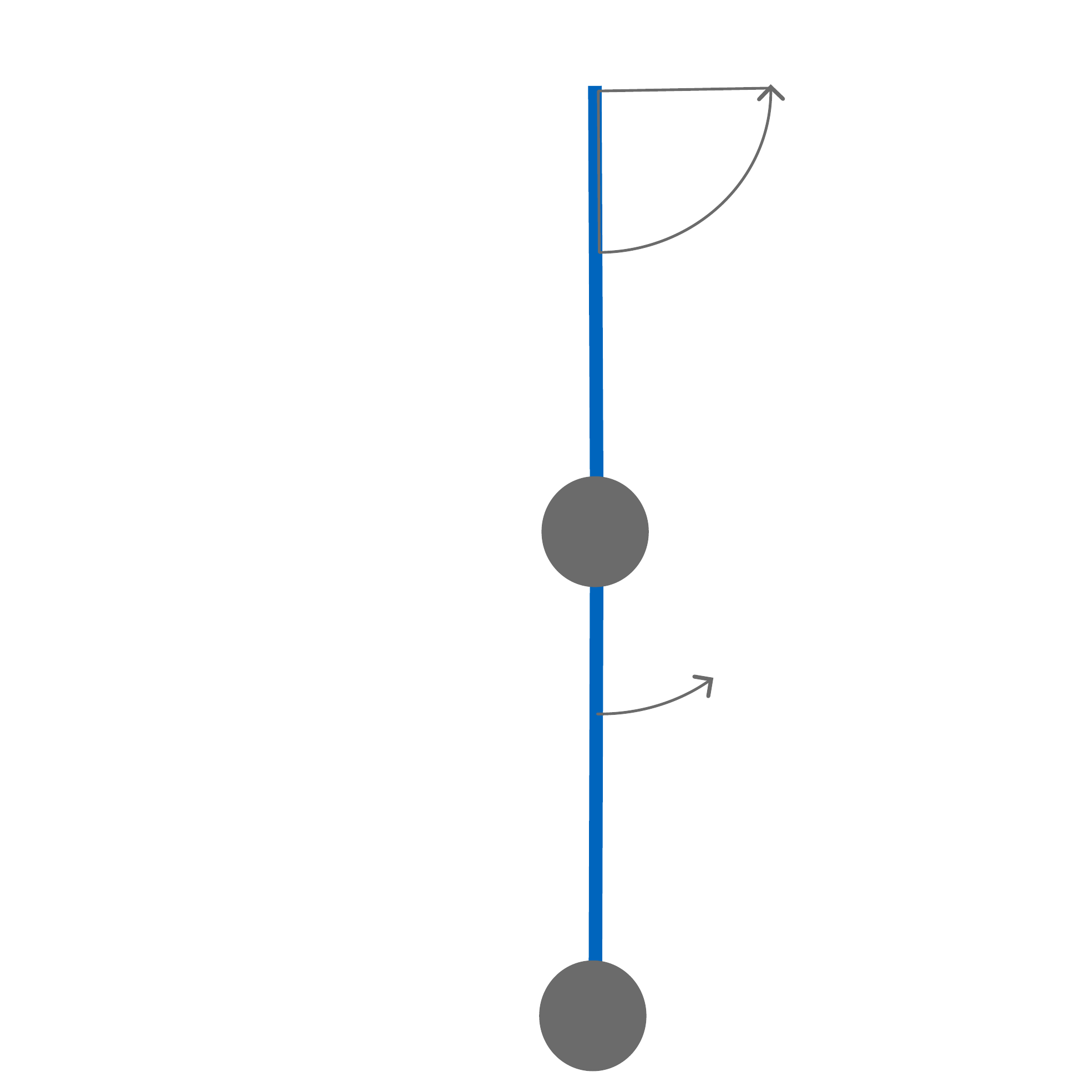}}%
    \put(0.55193155,0.85505341){\color[rgb]{0,0,0}\makebox(0,0)[lt]{\lineheight{1.25}\smash{\begin{tabular}[t]{l}$q_1$\end{tabular}}}}%
    \put(0.60192652,0.40228066){\color[rgb]{0,0,0}\makebox(0,0)[lt]{\lineheight{1.25}\smash{\begin{tabular}[t]{l}$q_2$\end{tabular}}}}%
    \put(0,0){\includegraphics[width=\unitlength,page=2]{dp_sym.pdf}}%
    \put(0.0883153,0.85321559){\color[rgb]{0,0,0}\makebox(0,0)[lt]{\lineheight{1.25}\smash{\begin{tabular}[t]{l}$g$\end{tabular}}}}%
  \end{picture}%
\endgroup%
}\hfill
  \subfloat[Some Periodic Trajectories\label{fig:class_on_torus}]{\def\svgwidth{.32\textwidth}\tiny%% Creator: Inkscape 1.2.2 (b0a8486541, 2022-12-01), www.inkscape.org
%% Accompanies image file 'torus_with_all.pdf' (pdf, eps, ps)
\begingroup%
  \makeatletter%
  \providecommand\color[2][]{%
    \errmessage{(Inkscape) Color is used for the text in Inkscape, but the package 'color.sty' is not loaded}%
    \renewcommand\color[2][]{}%
  }%
  \providecommand\transparent[1]{%
    \errmessage{(Inkscape) Transparency is used (non-zero) for the text in Inkscape, but the package 'transparent.sty' is not loaded}%
    \renewcommand\transparent[1]{}%
  }%
  \providecommand\rotatebox[2]{#2}%
  \newcommand*\fsize{\dimexpr\f@size pt\relax}%
  \newcommand*\lineheight[1]{\fontsize{\fsize}{#1\fsize}\selectfont}%
  \ifx\svgwidth\undefined%
    \setlength{\unitlength}{334.22390747bp}%
    \ifx\svgscale\undefined%
      \relax%
    \else%
      \setlength{\unitlength}{\unitlength * \real{\svgscale}}%
    \fi%
  \else%
    \setlength{\unitlength}{\svgwidth}%
  \fi%
  \global\let\svgwidth\undefined%
  \global\let\svgscale\undefined%
  \makeatother%
  \begin{picture}(1,0.86580266)%
    \lineheight{1}%
    \setlength\tabcolsep{0pt}%
    \put(0,0){\includegraphics[width=\unitlength,page=1]{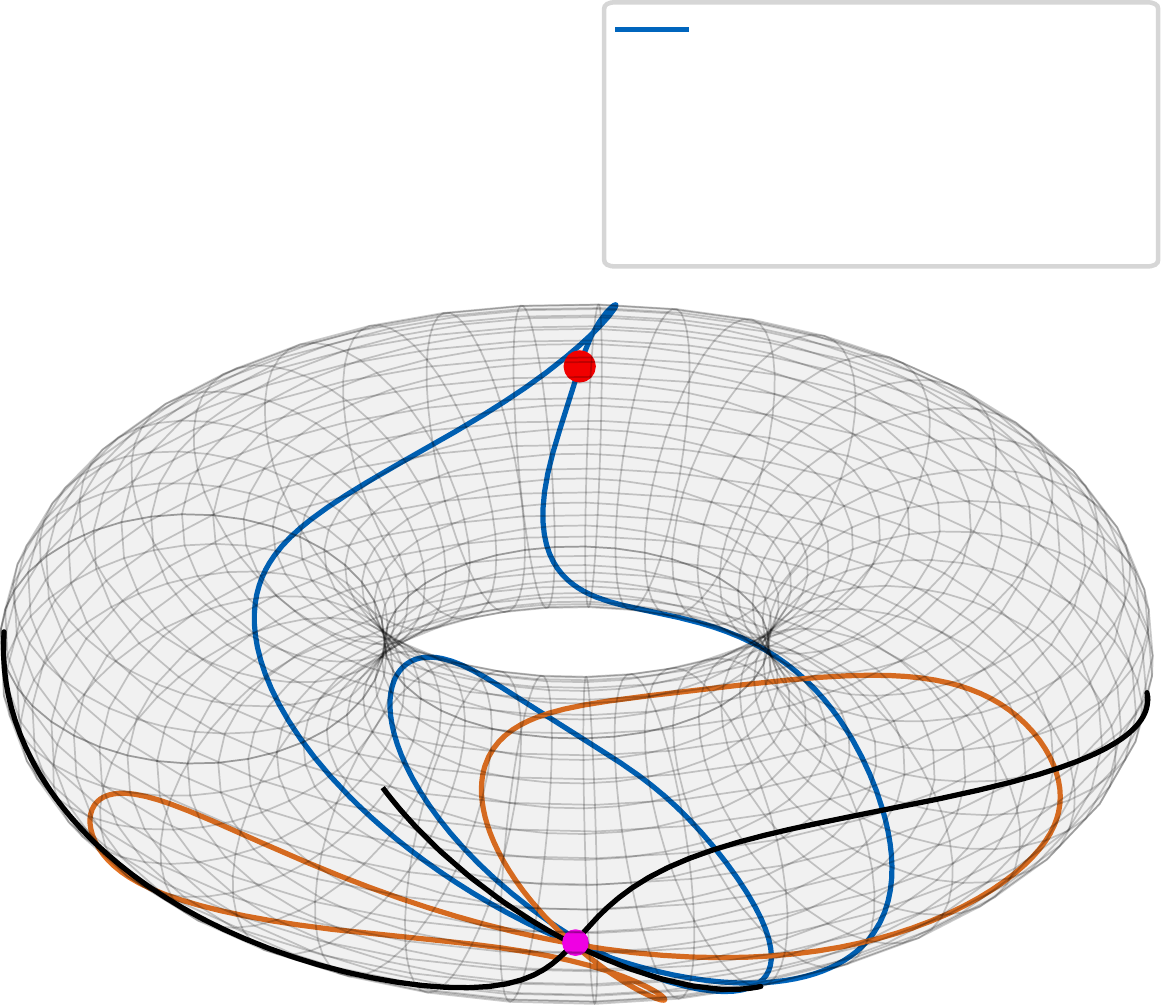}}%
    \put(0.61579906,0.82960406){\makebox(0,0)[lt]{\lineheight{0}\smash{\begin{tabular}[t]{l}Toroidal Orbits\end{tabular}}}}%
    \put(0,0){\includegraphics[width=\unitlength,page=2]{torus_with_all.pdf}}%
    \put(0.61579906,0.785687){\makebox(0,0)[lt]{\lineheight{0}\smash{\begin{tabular}[t]{l}Disk Orbit\end{tabular}}}}%
    \put(0,0){\includegraphics[width=\unitlength,page=3]{torus_with_all.pdf}}%
    \put(0.61579906,0.74176995){\makebox(0,0)[lt]{\lineheight{0}\smash{\begin{tabular}[t]{l}Brake Orbits\end{tabular}}}}%
    \put(0,0){\includegraphics[width=\unitlength,page=4]{torus_with_all.pdf}}%
    \put(0.61579906,0.69785289){\makebox(0,0)[lt]{\lineheight{0}\smash{\begin{tabular}[t]{l}$\arg\min U(q)$\end{tabular}}}}%
    \put(0,0){\includegraphics[width=\unitlength,page=5]{torus_with_all.pdf}}%
    \put(0.61579906,0.65393592){\makebox(0,0)[lt]{\lineheight{0}\smash{\begin{tabular}[t]{l}$\arg\max U(q)$\end{tabular}}}}%
  \end{picture}%
\endgroup%
}
  \caption{A double pendulum and some of its periodic trajectories of various classes shown on its toroidal configuration space.}
  \label{fig:dps}
\end{figure}

In this section we will discuss the large variety of periodic trajectories of conservative robot models based on one of the most simple examples, the 2-dof vertical robot, i.e. the double pendulum, see Fig.~\ref{fig:dps}.
We will classify the type of periodic orbits into three classes: toroidal orbits, disk orbits, and brake orbits.
Toroidal orbits are the ones directly predicted by algebraic topology; they are due to the toroidal structure of the configuration space.
These orbits turn at least one joint in full cycles. 
Disk orbits happen completely within a chart of disk topology, i.e., we do not need the wrapping of angles for them.
Finally, brake orbits are extensions of normal modes of linear systems.
Most types of trajectories discussed in this section will be present also for robots with arbitrary degrees of freedom.
Toroidal orbits are only possible as soon as one of the joints allows full turns.

\begin{figure*}[h!]
  \centering
  \begin{tabular}{rcc}
    $(0, 1)$ & \def\svgwidth{.65\textwidth}\tiny%% Creator: Inkscape 1.2.2 (b0a8486541, 2022-12-01), www.inkscape.org
%% Accompanies image file 'mn_01.pdf' (pdf, eps, ps)
\begingroup%
  \makeatletter%
  \providecommand\color[2][]{%
    \errmessage{(Inkscape) Color is used for the text in Inkscape, but the package 'color.sty' is not loaded}%
    \renewcommand\color[2][]{}%
  }%
  \providecommand\transparent[1]{%
    \errmessage{(Inkscape) Transparency is used (non-zero) for the text in Inkscape, but the package 'transparent.sty' is not loaded}%
    \renewcommand\transparent[1]{}%
  }%
  \providecommand\rotatebox[2]{#2}%
  \newcommand*\fsize{\dimexpr\f@size pt\relax}%
  \newcommand*\lineheight[1]{\fontsize{\fsize}{#1\fsize}\selectfont}%
  \ifx\svgwidth\undefined%
    \setlength{\unitlength}{508.35120013bp}%
    \ifx\svgscale\undefined%
      \relax%
    \else%
      \setlength{\unitlength}{\unitlength * \real{\svgscale}}%
    \fi%
  \else%
    \setlength{\unitlength}{\svgwidth}%
  \fi%
  \global\let\svgwidth\undefined%
  \global\let\svgscale\undefined%
  \makeatother%
  \begin{picture}(1,0.17830909)%
    \lineheight{1}%
    \setlength\tabcolsep{0pt}%
    \put(0,0){\includegraphics[width=\unitlength,page=1]{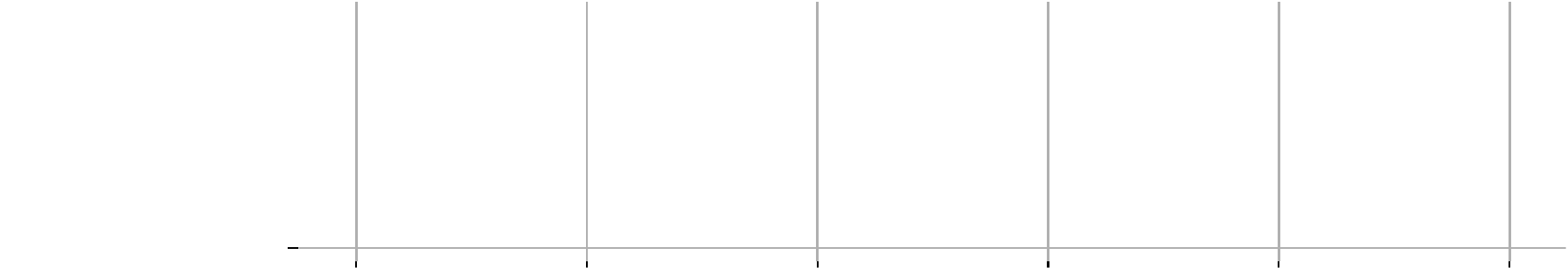}}%
    \put(0.17678107,0.0156026){\makebox(0,0)[rt]{\lineheight{0}\smash{\begin{tabular}[t]{r}$-\pi$\end{tabular}}}}%
    \put(0,0){\includegraphics[width=\unitlength,page=2]{mn_01.pdf}}%
    \put(0.17678107,0.08997633){\makebox(0,0)[rt]{\lineheight{0}\smash{\begin{tabular}[t]{r}0\end{tabular}}}}%
    \put(0,0){\includegraphics[width=\unitlength,page=3]{mn_01.pdf}}%
    \put(0.17678107,0.15844864){\makebox(0,0)[rt]{\lineheight{0}\smash{\begin{tabular}[t]{r}$\pi$\end{tabular}}}}%
    \put(0,0){\includegraphics[width=\unitlength,page=4]{mn_01.pdf}}%
    \put(0.93136138,0.14093224){\makebox(0,0)[lt]{\lineheight{1.25}\smash{\begin{tabular}[t]{l}$q_1$\end{tabular}}}}%
    \put(0,0){\includegraphics[width=\unitlength,page=5]{mn_01.pdf}}%
    \put(0.93136138,0.12342572){\makebox(0,0)[lt]{\lineheight{1.25}\smash{\begin{tabular}[t]{l}$q_2$\end{tabular}}}}%
    \put(0,0){\includegraphics[width=\unitlength,page=6]{mn_01.pdf}}%
  \end{picture}%
\endgroup%
 & \def\svgwidth{.25\textwidth}\tiny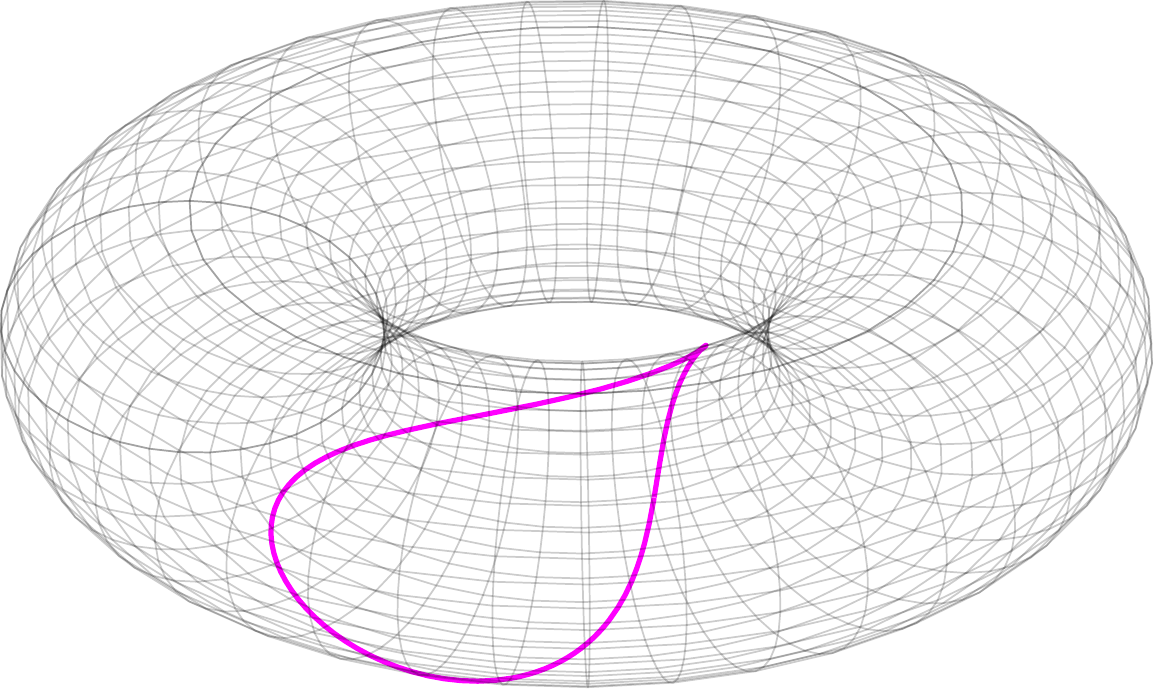\\
    $(1, 0)$ & \def\svgwidth{.65\textwidth}\tiny%% Creator: Inkscape 1.2.2 (b0a8486541, 2022-12-01), www.inkscape.org
%% Accompanies image file 'mn_10.pdf' (pdf, eps, ps)
\begingroup%
  \makeatletter%
  \providecommand\color[2][]{%
    \errmessage{(Inkscape) Color is used for the text in Inkscape, but the package 'color.sty' is not loaded}%
    \renewcommand\color[2][]{}%
  }%
  \providecommand\transparent[1]{%
    \errmessage{(Inkscape) Transparency is used (non-zero) for the text in Inkscape, but the package 'transparent.sty' is not loaded}%
    \renewcommand\transparent[1]{}%
  }%
  \providecommand\rotatebox[2]{#2}%
  \newcommand*\fsize{\dimexpr\f@size pt\relax}%
  \newcommand*\lineheight[1]{\fontsize{\fsize}{#1\fsize}\selectfont}%
  \ifx\svgwidth\undefined%
    \setlength{\unitlength}{497.49493456bp}%
    \ifx\svgscale\undefined%
      \relax%
    \else%
      \setlength{\unitlength}{\unitlength * \real{\svgscale}}%
    \fi%
  \else%
    \setlength{\unitlength}{\svgwidth}%
  \fi%
  \global\let\svgwidth\undefined%
  \global\let\svgscale\undefined%
  \makeatother%
  \begin{picture}(1,0.18915589)%
    \lineheight{1}%
    \setlength\tabcolsep{0pt}%
    \put(0,0){\includegraphics[width=\unitlength,page=1]{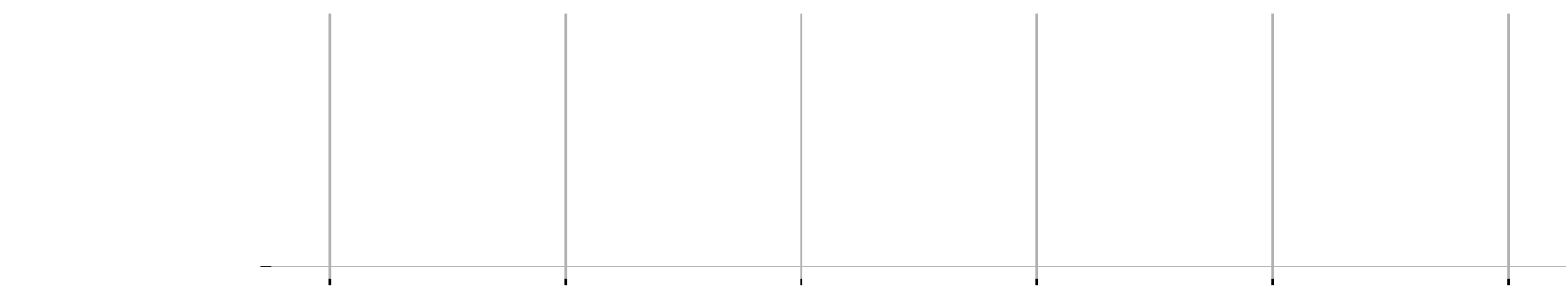}}%
    \put(0.15881692,0.0147166){\makebox(0,0)[rt]{\lineheight{0}\smash{\begin{tabular}[t]{r}$-\pi$\end{tabular}}}}%
    \put(0,0){\includegraphics[width=\unitlength,page=2]{mn_10.pdf}}%
    \put(0.15881692,0.0909228){\makebox(0,0)[rt]{\lineheight{0}\smash{\begin{tabular}[t]{r}0\end{tabular}}}}%
    \put(0,0){\includegraphics[width=\unitlength,page=3]{mn_10.pdf}}%
    \put(0.15881692,0.15319184){\makebox(0,0)[rt]{\lineheight{0}\smash{\begin{tabular}[t]{r}$\pi$\end{tabular}}}}%
    \put(0,0){\includegraphics[width=\unitlength,page=4]{mn_10.pdf}}%
  \end{picture}%
\endgroup%
 & \def\svgwidth{.25\textwidth}\tiny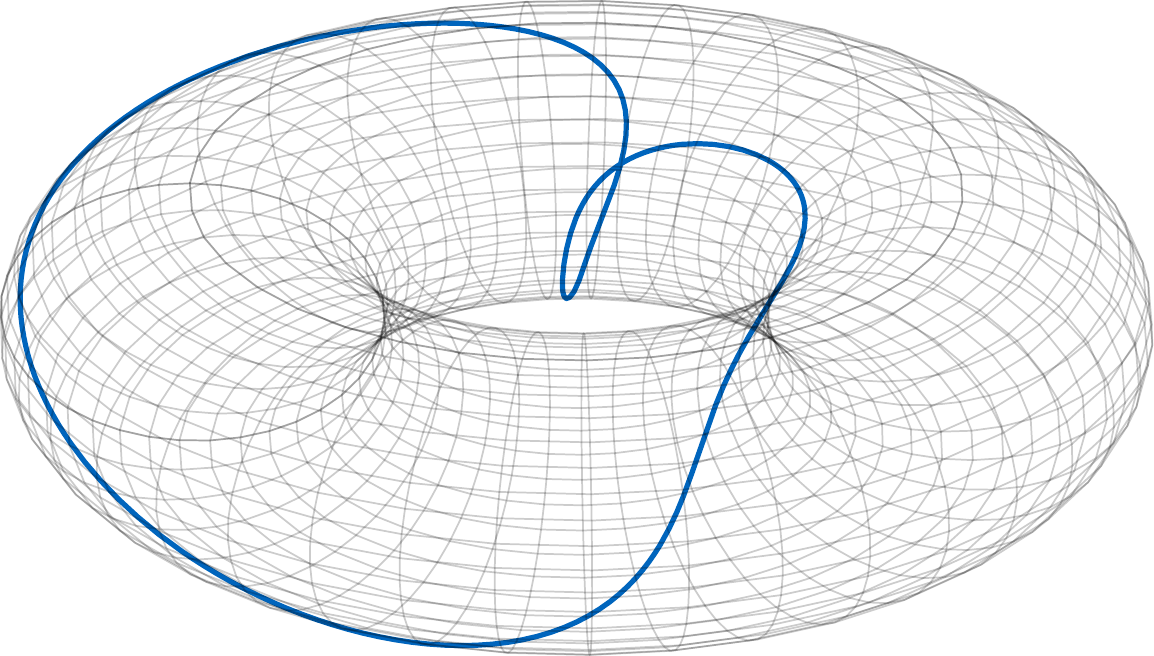\\
    $(1, 2)$ & \def\svgwidth{.65\textwidth}\tiny%% Creator: Inkscape 1.2.2 (b0a8486541, 2022-12-01), www.inkscape.org
%% Accompanies image file 'mn_12.pdf' (pdf, eps, ps)
\begingroup%
  \makeatletter%
  \providecommand\color[2][]{%
    \errmessage{(Inkscape) Color is used for the text in Inkscape, but the package 'color.sty' is not loaded}%
    \renewcommand\color[2][]{}%
  }%
  \providecommand\transparent[1]{%
    \errmessage{(Inkscape) Transparency is used (non-zero) for the text in Inkscape, but the package 'transparent.sty' is not loaded}%
    \renewcommand\transparent[1]{}%
  }%
  \providecommand\rotatebox[2]{#2}%
  \newcommand*\fsize{\dimexpr\f@size pt\relax}%
  \newcommand*\lineheight[1]{\fontsize{\fsize}{#1\fsize}\selectfont}%
  \ifx\svgwidth\undefined%
    \setlength{\unitlength}{498.00147349bp}%
    \ifx\svgscale\undefined%
      \relax%
    \else%
      \setlength{\unitlength}{\unitlength * \real{\svgscale}}%
    \fi%
  \else%
    \setlength{\unitlength}{\svgwidth}%
  \fi%
  \global\let\svgwidth\undefined%
  \global\let\svgscale\undefined%
  \makeatother%
  \begin{picture}(1,0.18608416)%
    \lineheight{1}%
    \setlength\tabcolsep{0pt}%
    \put(0,0){\includegraphics[width=\unitlength,page=1]{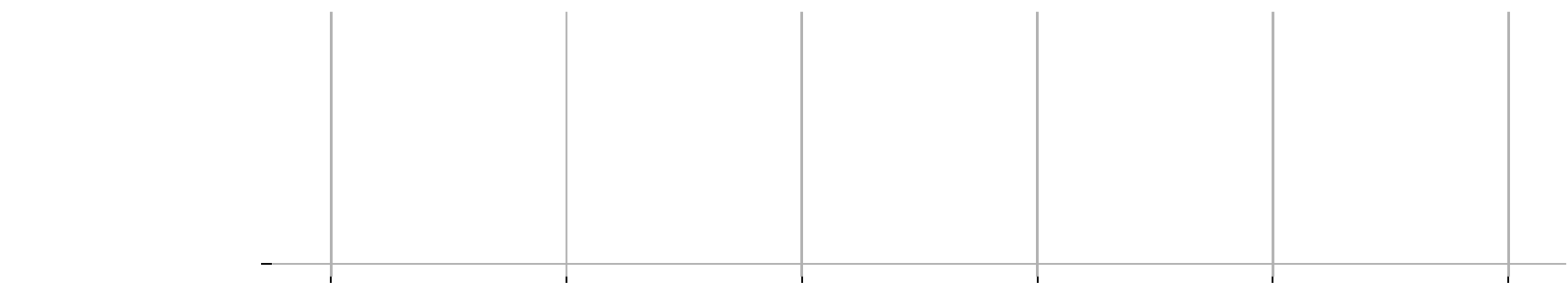}}%
    \put(0.15967253,0.01306878){\makebox(0,0)[rt]{\lineheight{0}\smash{\begin{tabular}[t]{r}$-\pi$\end{tabular}}}}%
    \put(0,0){\includegraphics[width=\unitlength,page=2]{mn_12.pdf}}%
    \put(0.15967253,0.08953148){\makebox(0,0)[rt]{\lineheight{0}\smash{\begin{tabular}[t]{r}0\end{tabular}}}}%
    \put(0,0){\includegraphics[width=\unitlength,page=3]{mn_12.pdf}}%
    \put(0.15967253,0.16412424){\makebox(0,0)[rt]{\lineheight{0}\smash{\begin{tabular}[t]{r}$\pi$\end{tabular}}}}%
    \put(0,0){\includegraphics[width=\unitlength,page=4]{mn_12.pdf}}%
  \end{picture}%
\endgroup%
 & \def\svgwidth{.25\textwidth}\tiny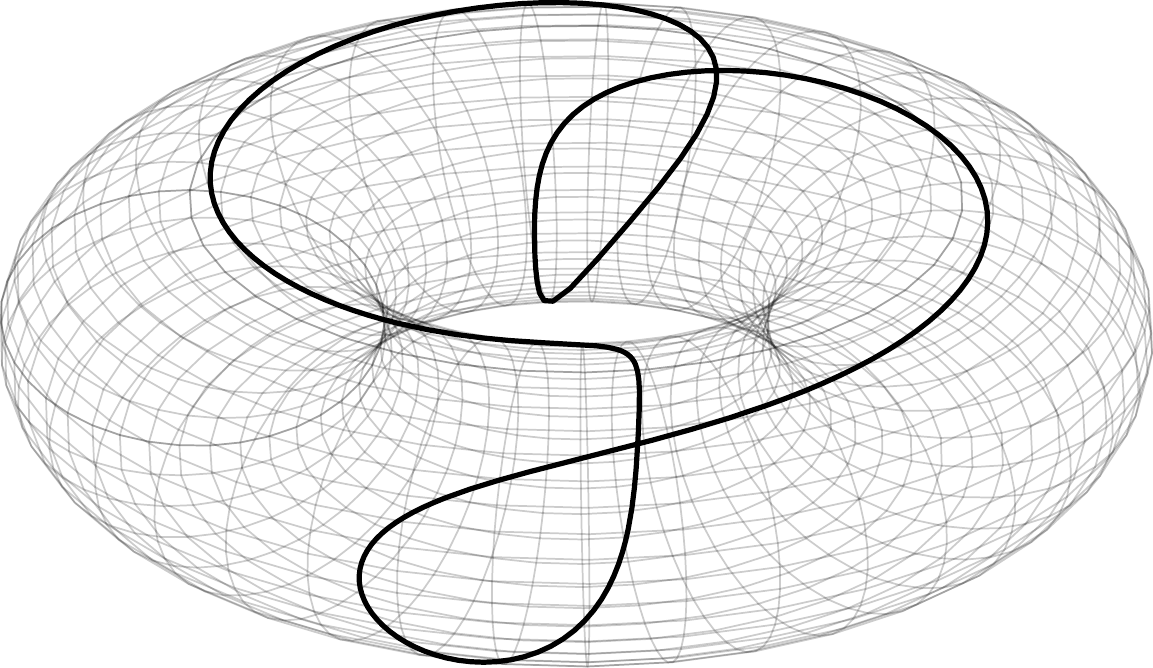\\
    $(2, 1)$ & \def\svgwidth{.65\textwidth}\tiny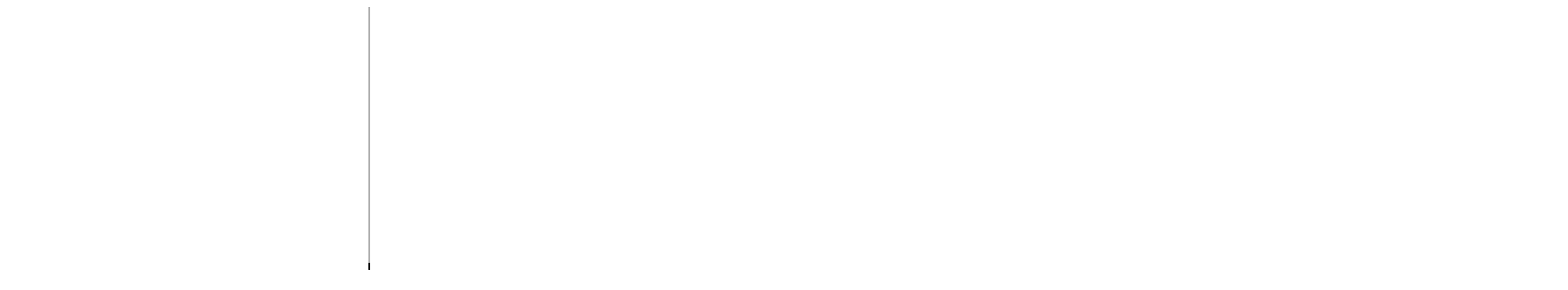 & \def\svgwidth{.25\textwidth}\tiny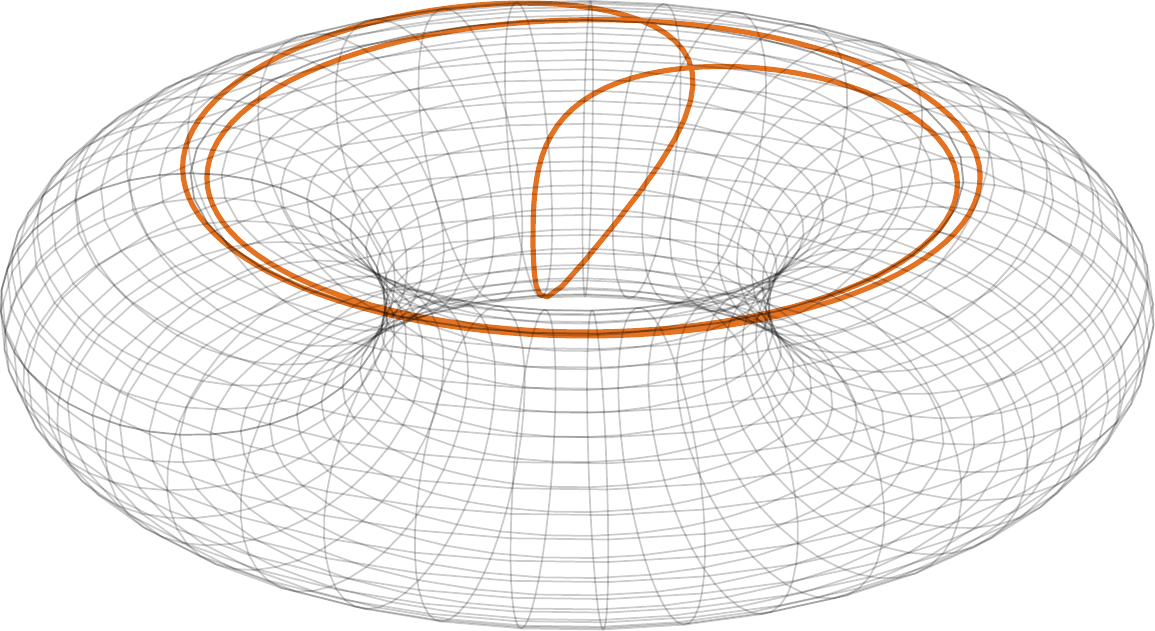\\
  \end{tabular}
  \caption{Toroidal orbits of the double pendulum for some fixed energies of types.
  $(0, 1)$, $(1, 0)$, and $(1, 2)$. On the left we show the Cartesian trajectories of the first (blue) and second (orange) points mass. In the middle: time evolutions of the joints. Right: configuration paths on the torus.}
  \label{tab:mn}
\end{figure*}

\subsection{Toroidal Orbits}\label{sec:m-n-orbits}
The dynamics of the conservative double pendulum in a gravity field is known to display chaotic behavior \cite{Shinbrot1992}.  
This is due to the inertial couplings and the upper bounded potential, having unstable equilibria at the upright configurations. If the total energy is high enough to permit full turns of the joints, i.e. $E> U_{max}(\qq)$, the system has, however, also infinitely many periodic orbits. This is a direct consequence of the topology of the torus and its homology class ${\cal{H}}_1(\mathbb{T}^2)$.
As shown in Sec.~\ref{sec:algebraic-topology}, there are infinitely many disjoint classes of closed curves, obtained by integer linear combination and homotopic deformation of the two base cycles on the torus. According to their definition, these curves cannot be shortened to a point, so there is a minimal length in each of these classes, and therefore the curve having that length will be a geodesic. 

We take the numerical string relaxation algorithm of Sec.~\ref{sec:curve shortening} to find such geodesics.
First, we fix the desired energy $E$ needed to determine the Jacobi metric (\ref{eq:jacobi_metric}).
Then we start by creating a string of the correct topology, i.e., we create an initial string in the class $(\alpha_1,\alpha_2)$ for $\alpha_1, \alpha_2 \in \mathbb{Z}$.
Iteratively updating the nodes of the string by the update rule will let the string converge to a geodesic - and thus to a periodic orbit of the double pendulum.
This result holds in any dimension, as the Betti number for an n-dof pendulum is $b_1(\mathbb{T}^n)=n$. It is indeed a classical result \cite{Arnold1989} that
\begin{proposition}
For any n integers $\alpha_1,...\alpha_n$, of which at least one is nonzero, there exists at least one periodic trajectory of the n-dof pendulum performing $\alpha_i$ rotations around joint $i$, for $i=1,...,n$.
\end{proposition}

Fig.~\ref{tab:mn} shows the trajectories of the double pendulum in the gravity field for the combinations $(0,1)$, $(1,0)$, $(1,2)$, and $(2,1)$ for some fixed energies.
It is important to note that each orbit is only valid for the energy it was computed for; the geodesics are not invariant w.r.t. the energy in the Jacobi metric.
We observe, however, that they continuously deform with variations in energy.
For the double pendulum, we find sometimes more than one geodesic, also for fixed energies.
Only one of them will be a global minimum, others only local ones.

\subsection{Disk Orbits}
The algebraic topology results do not say anything about the existence of periodic orbits of the type $(0,0)$, i.e., of closed trajectories that oscillate in an interval less than $2\pi$. Indeed, such trajectories do not need to exist in general (for arbitrary metrics), because all the zero cycles can be shrunk to a point; there is no hole to prevent their collapse. 
Nevertheless, it is not excluded that the metric encodes \emph{bumps} or other local geometrical features such that closed curves make the arc length stationary: this makes them geodesics and, simultaneously, periodic orbits.
Indeed, studies on chaotic systems show that they often display a rich variety of unstable periodic orbits~\cite{Saiki2007,Hogan1999}.
This has also been shown for mechanical systems \cite{Jahn2021,Phipps2006}.
We employ a scheme similar to \cite{Saiki2007}: we take a boundary value solver \cite{Ascher1995} to find solutions $\q(t)$ to the differential equation (\ref{eq:euler_lagrange_eq}) such that $\q(0) = \q(T)$, $\dqq(0) = \dqq(T)$ and $H(\q, \dqq) = E_{\text{des}}$ for some desired energy $E_{\text{des}}$.
The estimated period time $T$ is updated as well during the optimization.
Once we found a solution we perform numerical continuation \cite{Dankowicz2013} over the energy to generate families of solutions.

\begin{figure*}
  \centering
  \subfloat{\def\svgwidth{.3\textwidth}\tiny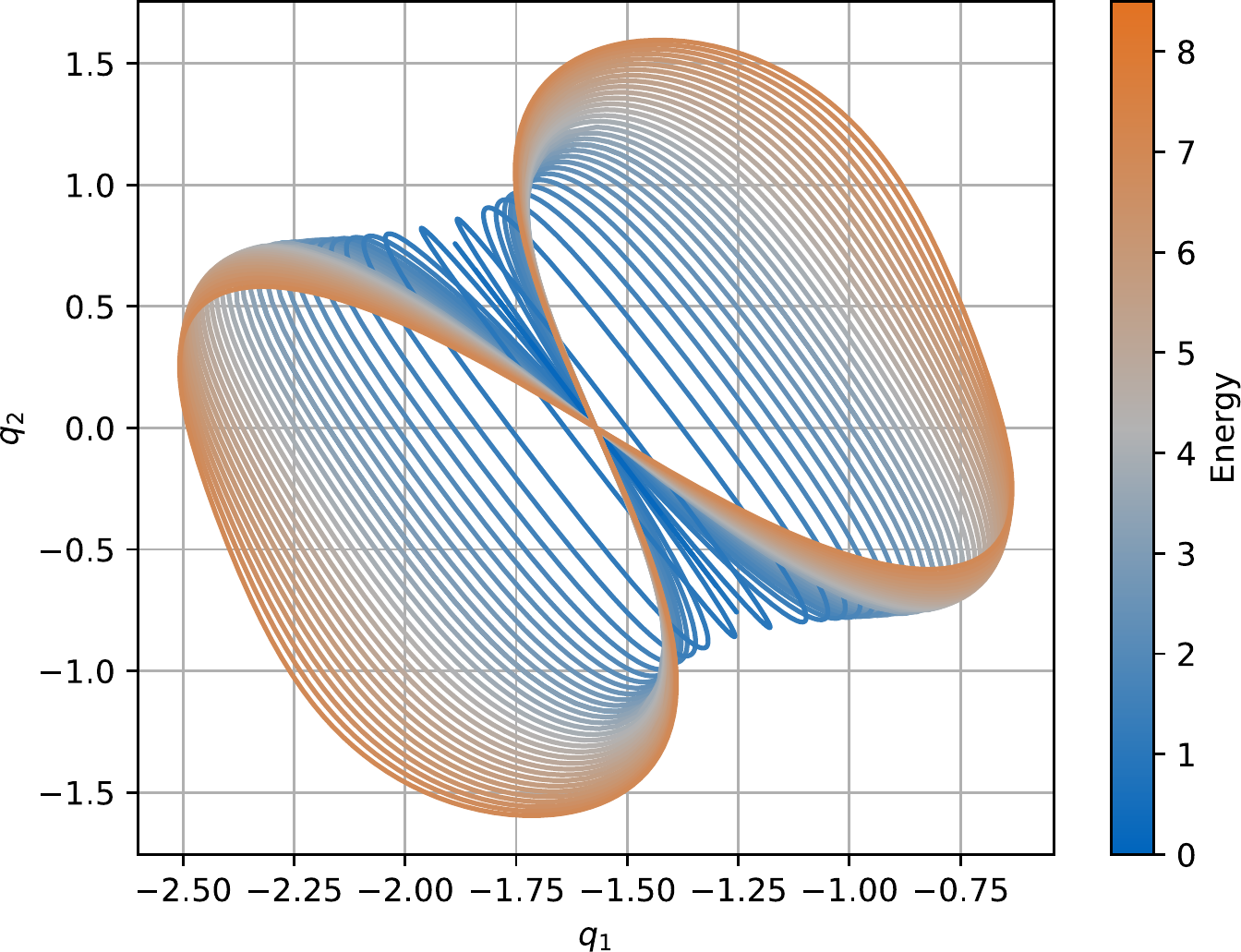}\hspace{.5cm}
  \subfloat{\def\svgwidth{.2\textwidth}\tiny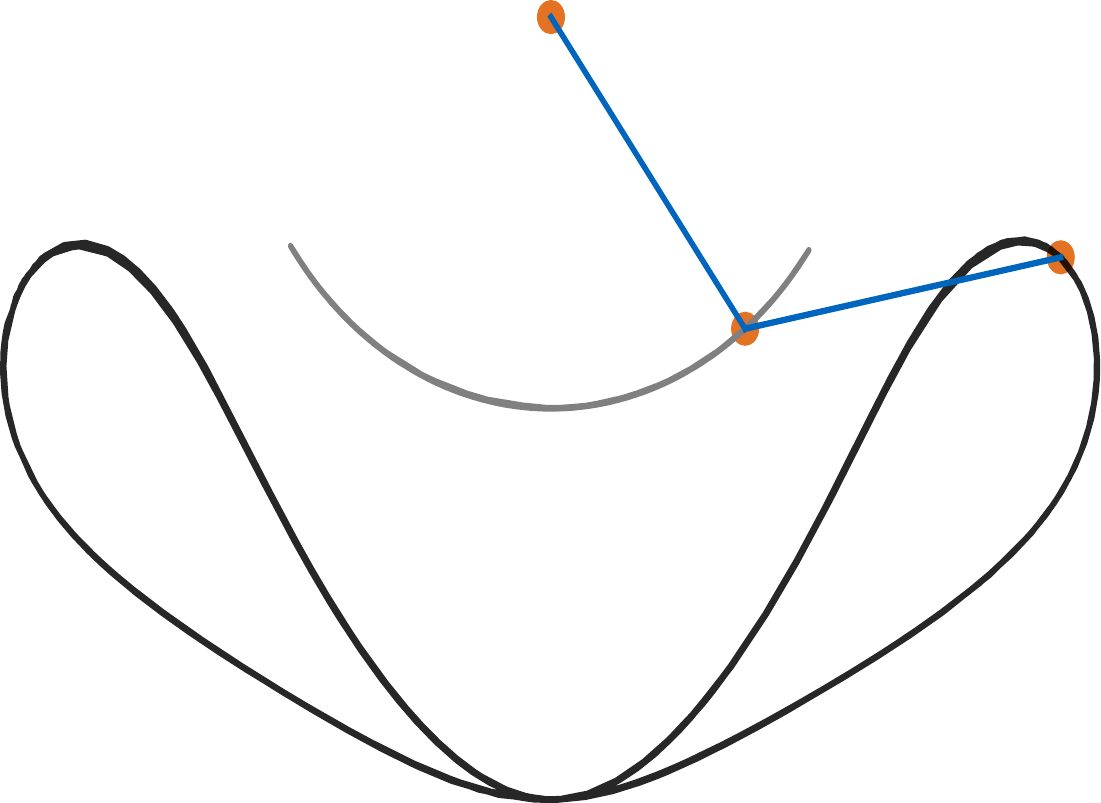}\hspace{.5cm}
  \subfloat{\def\svgwidth{.3\textwidth}\tiny%% Creator: Inkscape 1.2.2 (b0a8486541, 2022-12-01), www.inkscape.org
%% Accompanies image file '00_eight_q.pdf' (pdf, eps, ps)
\begingroup%
  \makeatletter%
  \providecommand\color[2][]{%
    \errmessage{(Inkscape) Color is used for the text in Inkscape, but the package 'color.sty' is not loaded}%
    \renewcommand\color[2][]{}%
  }%
  \providecommand\transparent[1]{%
    \errmessage{(Inkscape) Transparency is used (non-zero) for the text in Inkscape, but the package 'transparent.sty' is not loaded}%
    \renewcommand\transparent[1]{}%
  }%
  \providecommand\rotatebox[2]{#2}%
  \newcommand*\fsize{\dimexpr\f@size pt\relax}%
  \newcommand*\lineheight[1]{\fontsize{\fsize}{#1\fsize}\selectfont}%
  \ifx\svgwidth\undefined%
    \setlength{\unitlength}{438.49530029bp}%
    \ifx\svgscale\undefined%
      \relax%
    \else%
      \setlength{\unitlength}{\unitlength * \real{\svgscale}}%
    \fi%
  \else%
    \setlength{\unitlength}{\svgwidth}%
  \fi%
  \global\let\svgwidth\undefined%
  \global\let\svgscale\undefined%
  \makeatother%
  \begin{picture}(1,0.73446771)%
    \lineheight{1}%
    \setlength\tabcolsep{0pt}%
    \put(0,0){\includegraphics[width=\unitlength,page=1]{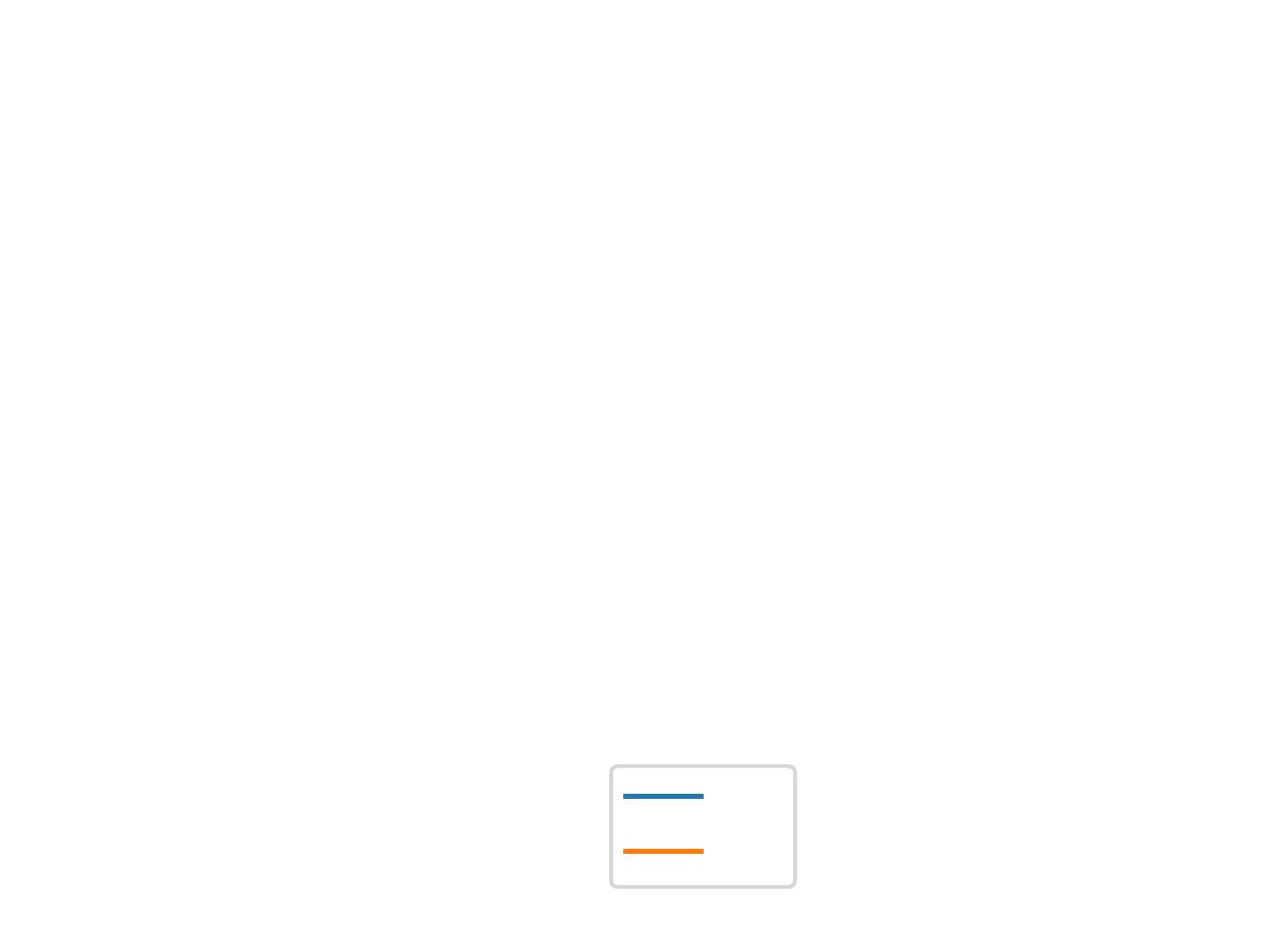}}%
    \put(0.5533825,0.10936664){\color[rgb]{0,0,0}\makebox(0,0)[lt]{\lineheight{1.25}\smash{\begin{tabular}[t]{l}$\dot{q}_1$\end{tabular}}}}%
    \put(0.55262534,0.06568999){\color[rgb]{0,0,0}\makebox(0,0)[lt]{\lineheight{1.25}\smash{\begin{tabular}[t]{l}$\dot{q}_2$\end{tabular}}}}%
    \put(0,0){\includegraphics[width=\unitlength,page=2]{00_eight_q.pdf}}%
    \put(0.91804798,0.68332373){\color[rgb]{0,0,0}\makebox(0,0)[lt]{\lineheight{1.25}\smash{\begin{tabular}[t]{l}$q_1$\end{tabular}}}}%
    \put(0.91729082,0.63964709){\color[rgb]{0,0,0}\makebox(0,0)[lt]{\lineheight{1.25}\smash{\begin{tabular}[t]{l}$q_2$\end{tabular}}}}%
    \put(0.48471408,-0.03458963){\color[rgb]{0,0,0}\makebox(0,0)[lt]{\lineheight{1.25}\smash{\begin{tabular}[t]{l}Time\end{tabular}}}}%
  \end{picture}%
\endgroup%
}\\

  \subfloat{\def\svgwidth{.3\textwidth}\tiny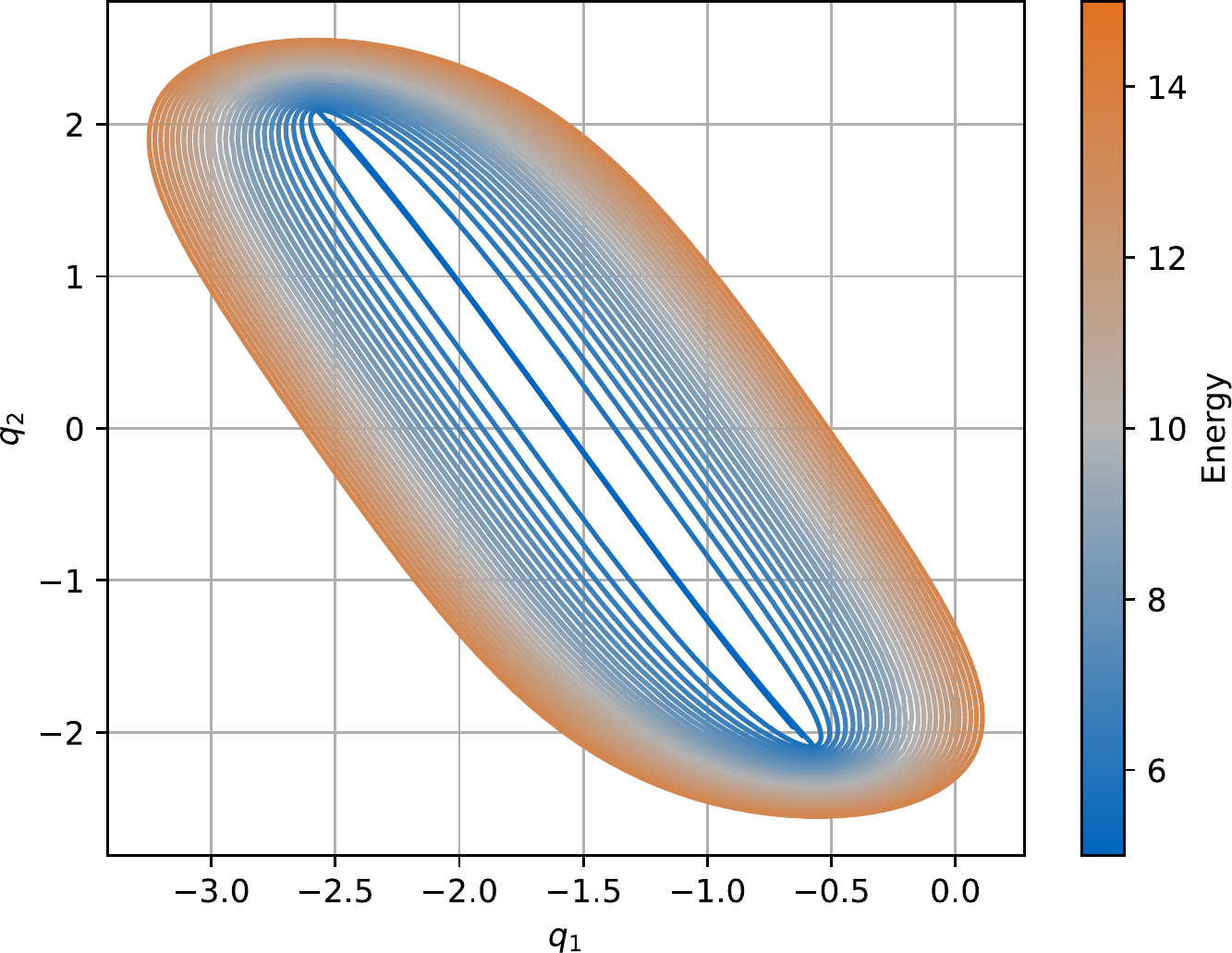}\hspace{.5cm}
  \subfloat{\def\svgwidth{.2\textwidth}\tiny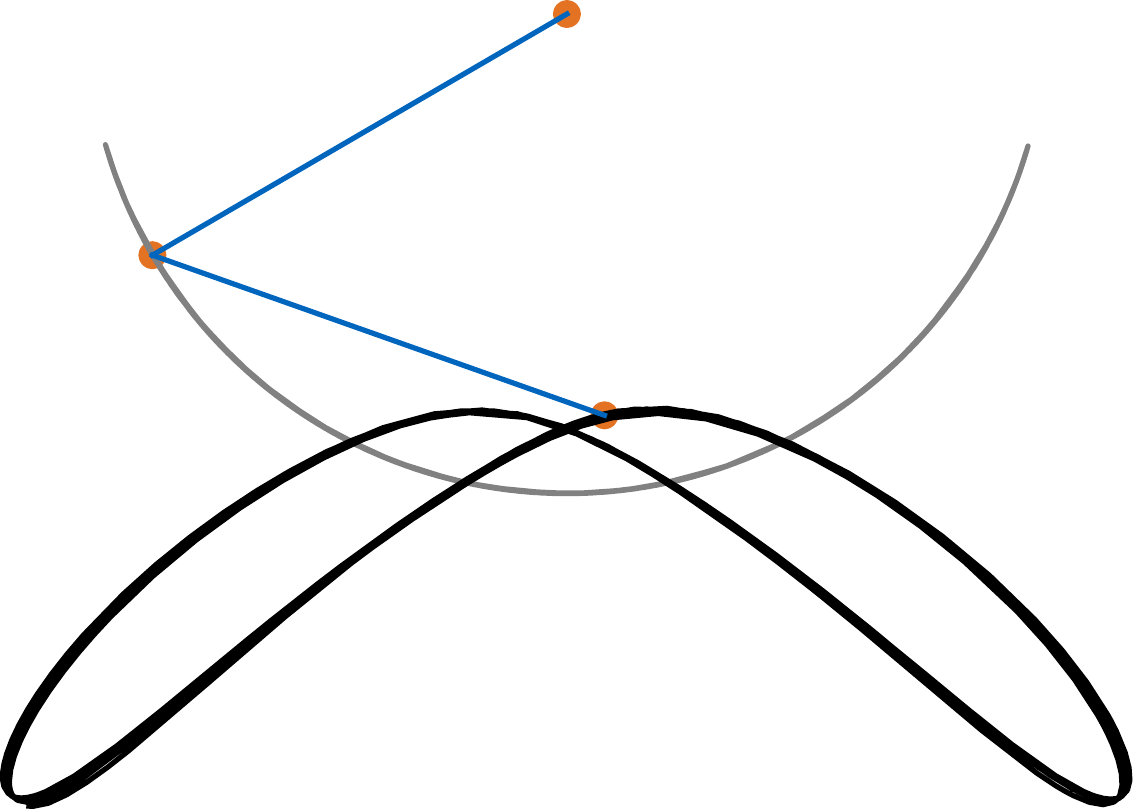}\hspace{.5cm}
  \subfloat{\def\svgwidth{.3\textwidth}\tiny%% Creator: Inkscape 1.2.2 (b0a8486541, 2022-12-01), www.inkscape.org
%% Accompanies image file '00_zero_q.pdf' (pdf, eps, ps)
\begingroup%
  \makeatletter%
  \providecommand\color[2][]{%
    \errmessage{(Inkscape) Color is used for the text in Inkscape, but the package 'color.sty' is not loaded}%
    \renewcommand\color[2][]{}%
  }%
  \providecommand\transparent[1]{%
    \errmessage{(Inkscape) Transparency is used (non-zero) for the text in Inkscape, but the package 'transparent.sty' is not loaded}%
    \renewcommand\transparent[1]{}%
  }%
  \providecommand\rotatebox[2]{#2}%
  \newcommand*\fsize{\dimexpr\f@size pt\relax}%
  \newcommand*\lineheight[1]{\fontsize{\fsize}{#1\fsize}\selectfont}%
  \ifx\svgwidth\undefined%
    \setlength{\unitlength}{438.49530029bp}%
    \ifx\svgscale\undefined%
      \relax%
    \else%
      \setlength{\unitlength}{\unitlength * \real{\svgscale}}%
    \fi%
  \else%
    \setlength{\unitlength}{\svgwidth}%
  \fi%
  \global\let\svgwidth\undefined%
  \global\let\svgscale\undefined%
  \makeatother%
  \begin{picture}(1,0.73446771)%
    \lineheight{1}%
    \setlength\tabcolsep{0pt}%
    \put(0,0){\includegraphics[width=\unitlength,page=1]{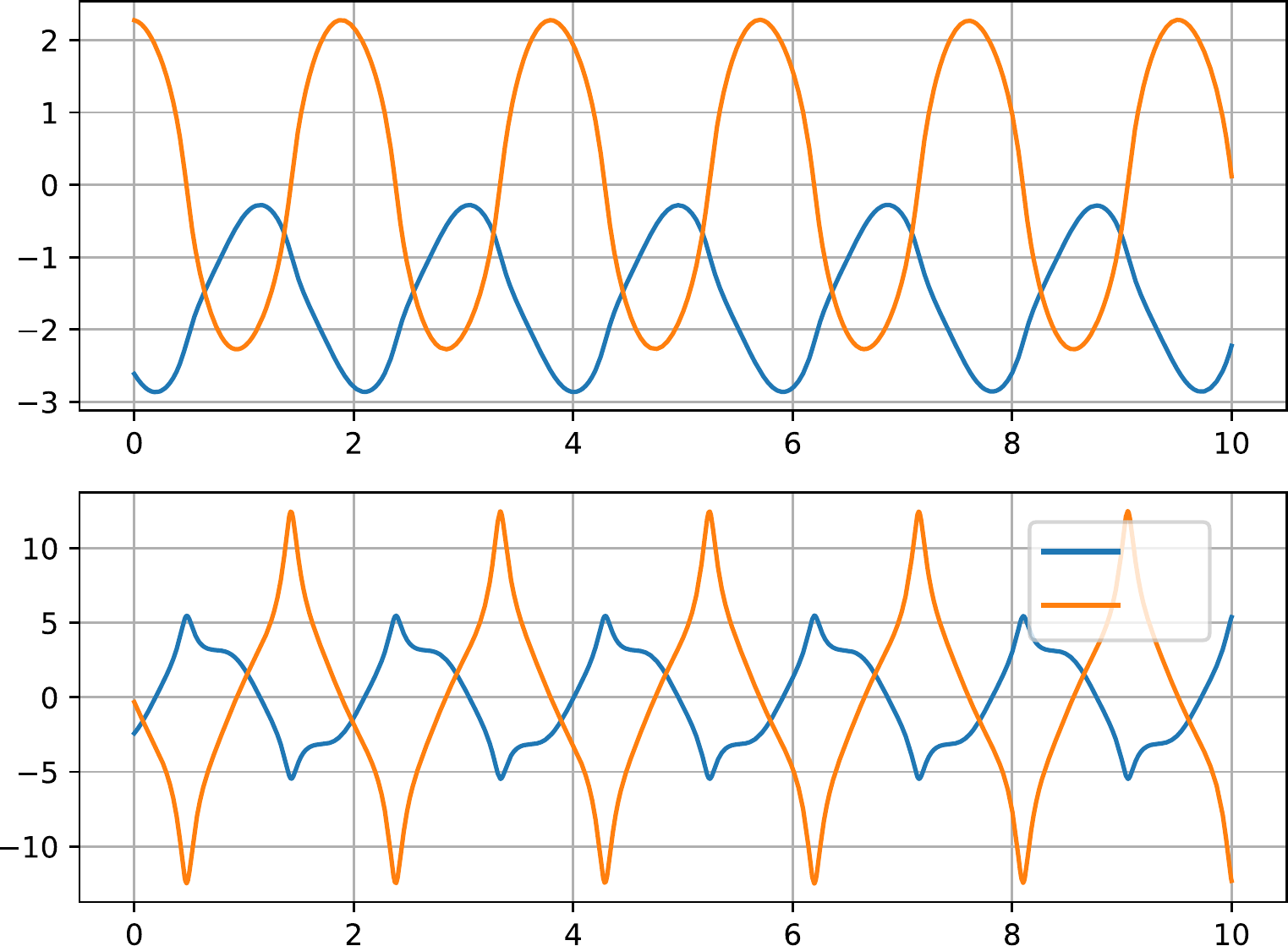}}%
    \put(0.88209323,0.296097){\color[rgb]{0,0,0}\makebox(0,0)[lt]{\lineheight{1.25}\smash{\begin{tabular}[t]{l}$\dot{q}_1$\end{tabular}}}}%
    \put(0.88135166,0.25331994){\color[rgb]{0,0,0}\makebox(0,0)[lt]{\lineheight{1.25}\smash{\begin{tabular}[t]{l}$\dot{q}_2$\end{tabular}}}}%
    \put(0,0){\includegraphics[width=\unitlength,page=2]{00_zero_q.pdf}}%
    \put(0.84950208,0.66293933){\color[rgb]{0,0,0}\makebox(0,0)[lt]{\lineheight{1.25}\smash{\begin{tabular}[t]{l}$q_1$\end{tabular}}}}%
    \put(0.8487135,0.61744991){\color[rgb]{0,0,0}\makebox(0,0)[lt]{\lineheight{1.25}\smash{\begin{tabular}[t]{l}$q_2$\end{tabular}}}}%
    \put(0.49518767,-0.03646789){\color[rgb]{0,0,0}\makebox(0,0)[lt]{\lineheight{1.25}\smash{\begin{tabular}[t]{l}Time\end{tabular}}}}%
  \end{picture}%
\endgroup%
}
  \caption{Disk Orbits for two different families. All angles in rad.}
  \label{fig:restless}
\end{figure*}

We have found a large variety of periodic orbits and show two particularly simple families in Fig.~\ref{fig:restless}, which continuously vary with energy.
Each row in Fig.~\ref{fig:restless} shows one family.
On the left we show the orbits in configuration space and observe that they continuously deform with energy.
On the middle we show the highest energy orbits in Cartesian space; and on the right the same orbit in configuration space.
It looks like, for low energies, they collapse into brake orbits (more to those later in Sec.~\ref{sec:brake_orbits}), although this observation has no theoretical backup yet. 
It will be certainly interesting to further investigate these types of trajectories in the future, because they are well suited for robots and also for biological limbs, which cannot perform full turns. For example, the swinging motion of a leg could be performed in such a mode.

\subsection{Brake Orbits aka Nonlinear Normal Modes}\label{sec:brake_orbits}
Brake orbits were the primary focus of our initial research interest and were presented in detail in~\cite{AlbuSchaeffer2020,AlbuSchaeffer2021}.
We understand this type of orbits as generalization of normal modes of linear systems.
For linear dynamics, periodic motions will take place in configuration space in the directions given by the eigenvectors. 
A line of research~\cite{Kerschen2009,Shaw1993} dating back to Rosenberg~\cite{Rosenberg1966} noted that there is a straightforward generalization to nonlinear systems and therefore called these oscillations {\emph{nonlinear normal modes}}. 
These modes were studied, however, only for systems composed of point masses and nonlinear potentials, thus not being applicable to robotic systems. 
In order to extend the results to robotics and to emphasize the connection to the linear modes, we coined the concept of {\emph{eigenmanifolds}} in~\cite{AlbuSchaeffer2020}. 
Each trajectory has the property that it oscillates back and forth between two points, where the system stops and reverses motion.

It turns out that trajectories of this kind have also been studied in a quite general setup with tools of differential geometry and algebraic topology since Seifert~\cite{Seifert1948}.
He has proven the existance of one brake orbit and conjectured there should be even more, both in his seminal work~\cite{Seifert1948}:
\begin{proposition}
For any conservative mechanical systems with closed equipotential surfaces, there exists at least one brake orbit for each energy level.
\end{proposition}
\begin{conjecture}[Seifert]
For any conservative mechanical systems with $n$ degrees of freedom and with closed equipotential surfaces, there exist at least $n$ brake orbits for each energy level.
\end{conjecture}
This is an idea certainly inspired by the $n$ modes of $n$ DoF linear systems. 
Some authors have provided proofs of the conjecture for particular cases \cite{Gluck1983} and Giambó et al.\ claim in a recent preprint to have proven it \cite{Giambo2022} under conditions which apply to general Hamiltonian systems.
However, the additional conditions of the theorems are generally not satisfied by robot dynamics equations, leaving this as a still open theoretical question.

\begin{figure*}
  \centering
  \subfloat{\def\svgwidth{.42\textwidth}\tiny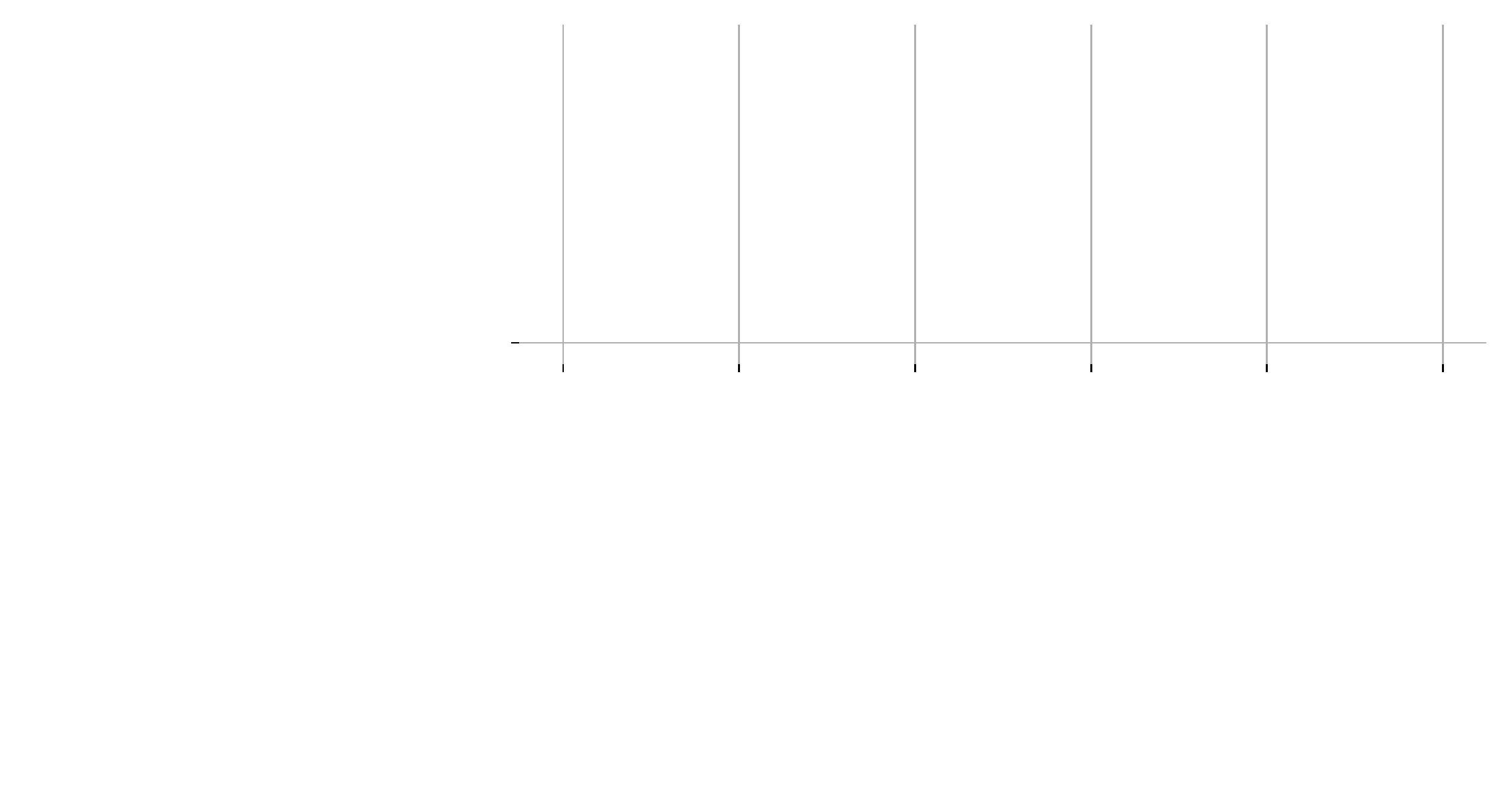}\hfill
  \subfloat{\def\svgwidth{.29\textwidth}\tiny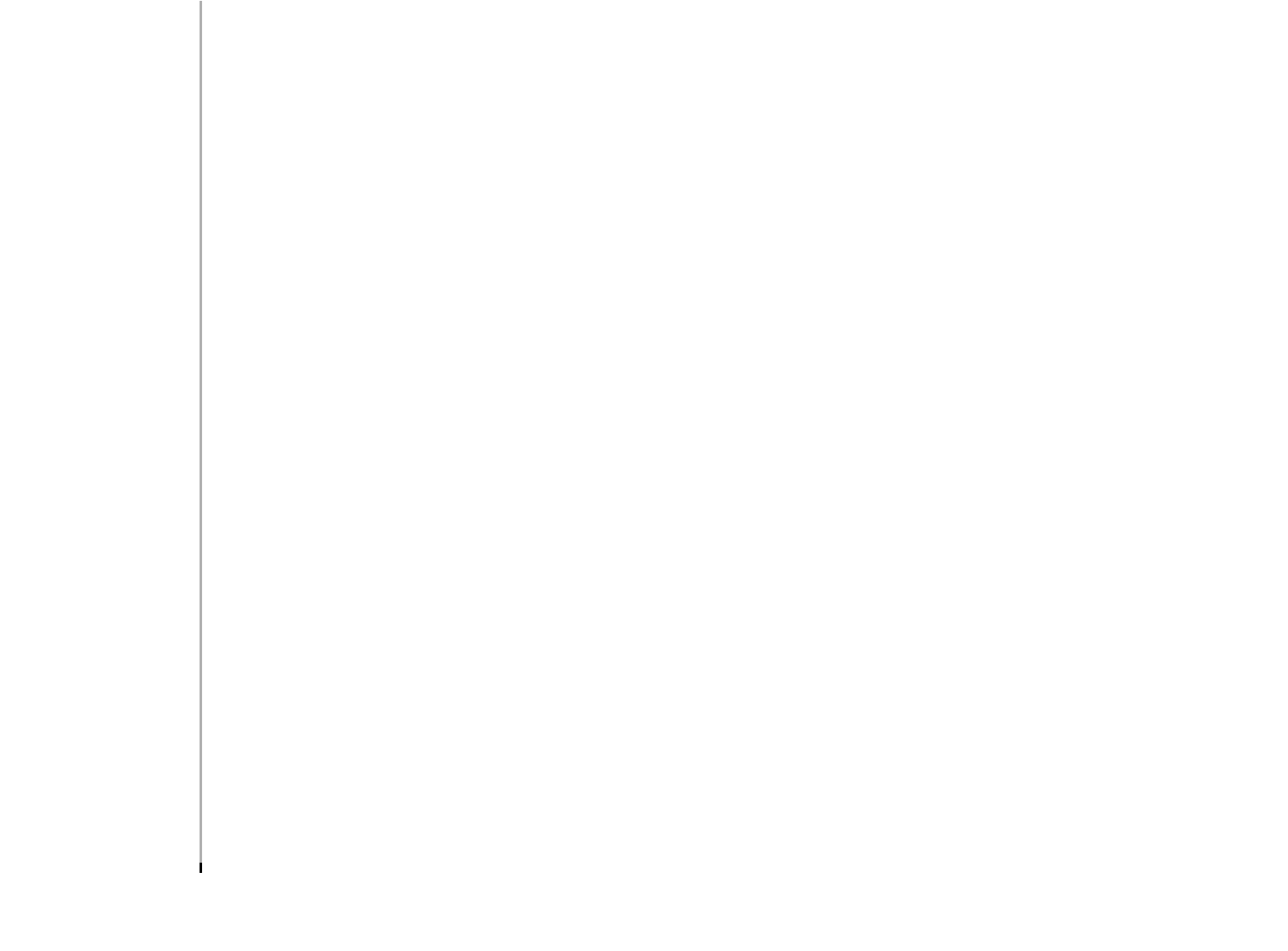}\hfill
  \subfloat{\def\svgwidth{.27\textwidth}\tiny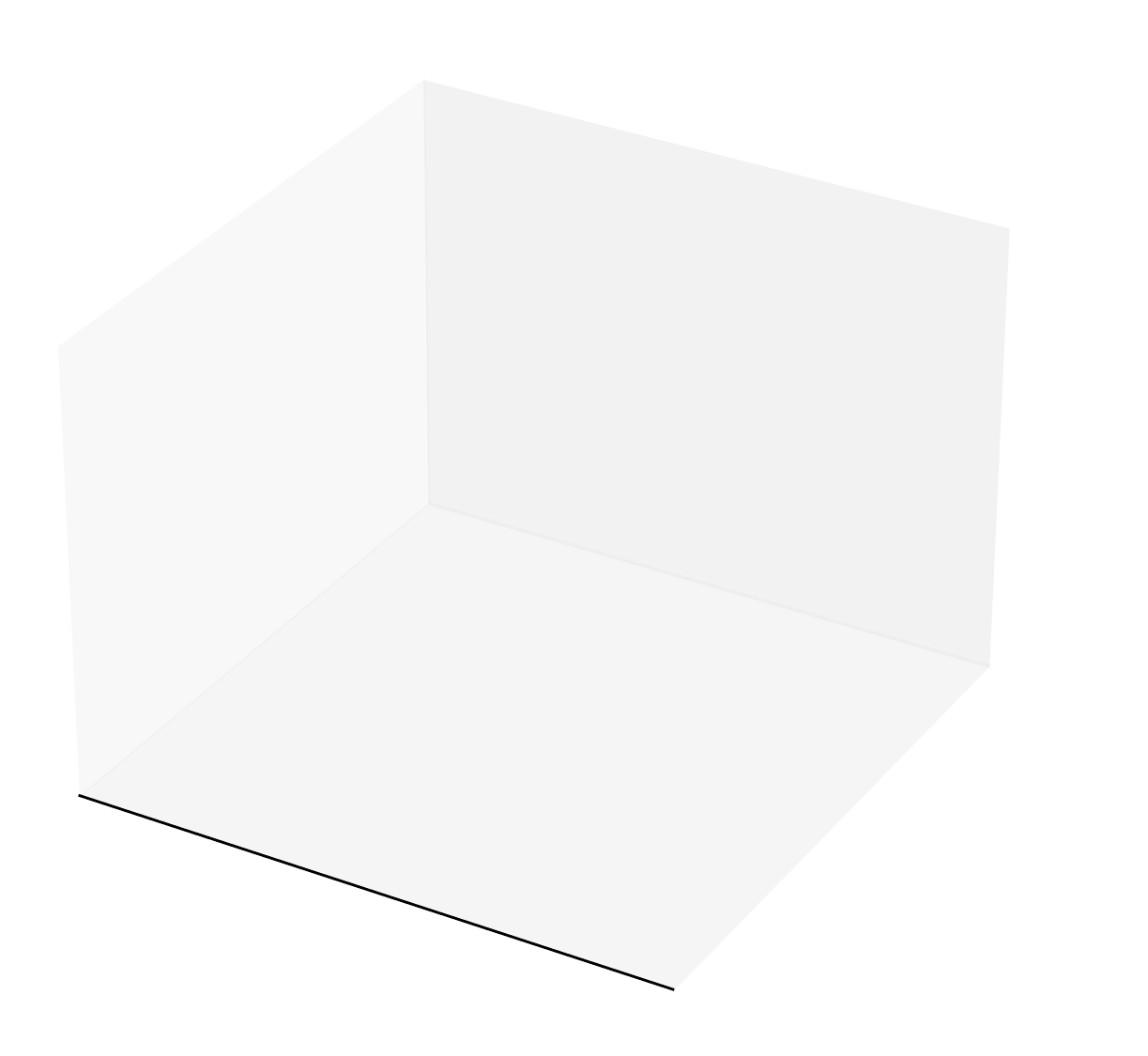}
  \caption{Brake orbit example for the double pendulum.}
  \label{fig:break_orbits}
\end{figure*}

Based on the insights of the theory, we developed numerical algorithms for searching the nonlinear modes (brake orbits) starting from the linearized solutions. 
Fig.~\ref{fig:break_orbits} presents brake orbits of the double pendulum.
On the left we show brake orbits growing our of the two linear eigenvectors.
The dots show two configurations, which are used for initial conditions for the simulations on the right.
In the middle, we display the brake orbits as a surface parameterized by energy as additional coordinate. 
This is a representation of the eigenmanifolds, which is alternative to the one from~\cite{AlbuSchaeffer2020}.
As predicted by the theorem, at least two nonlinear modes exist for every energy level.
We report that for all robot systems analyzed so far, including legged robots and a 7 DoF robot arm~\cite{Bjelonic2022}, the Seifert conjecture holds.

\subsection{Classification and Overview on Periodic Orbits}
\begin{figure}
  \centering
  \def\svgwidth{.5\textwidth}
  \scriptsize%% Creator: Inkscape 1.2.2 (b0a8486541, 2022-12-01), www.inkscape.org
%% Accompanies image file 'orbits.pdf' (pdf, eps, ps)
\begingroup%
  \makeatletter%
  \providecommand\color[2][]{%
    \errmessage{(Inkscape) Color is used for the text in Inkscape, but the package 'color.sty' is not loaded}%
    \renewcommand\color[2][]{}%
  }%
  \providecommand\transparent[1]{%
    \errmessage{(Inkscape) Transparency is used (non-zero) for the text in Inkscape, but the package 'transparent.sty' is not loaded}%
    \renewcommand\transparent[1]{}%
  }%
  \providecommand\rotatebox[2]{#2}%
  \newcommand*\fsize{\dimexpr\f@size pt\relax}%
  \newcommand*\lineheight[1]{\fontsize{\fsize}{#1\fsize}\selectfont}%
  \ifx\svgwidth\undefined%
    \setlength{\unitlength}{969.28152466bp}%
    \ifx\svgscale\undefined%
      \relax%
    \else%
      \setlength{\unitlength}{\unitlength * \real{\svgscale}}%
    \fi%
  \else%
    \setlength{\unitlength}{\svgwidth}%
  \fi%
  \global\let\svgwidth\undefined%
  \global\let\svgscale\undefined%
  \makeatother%
  \begin{picture}(1,0.32802207)%
    \lineheight{1}%
    \setlength\tabcolsep{0pt}%
    \put(0,0){\includegraphics[width=\unitlength,page=1]{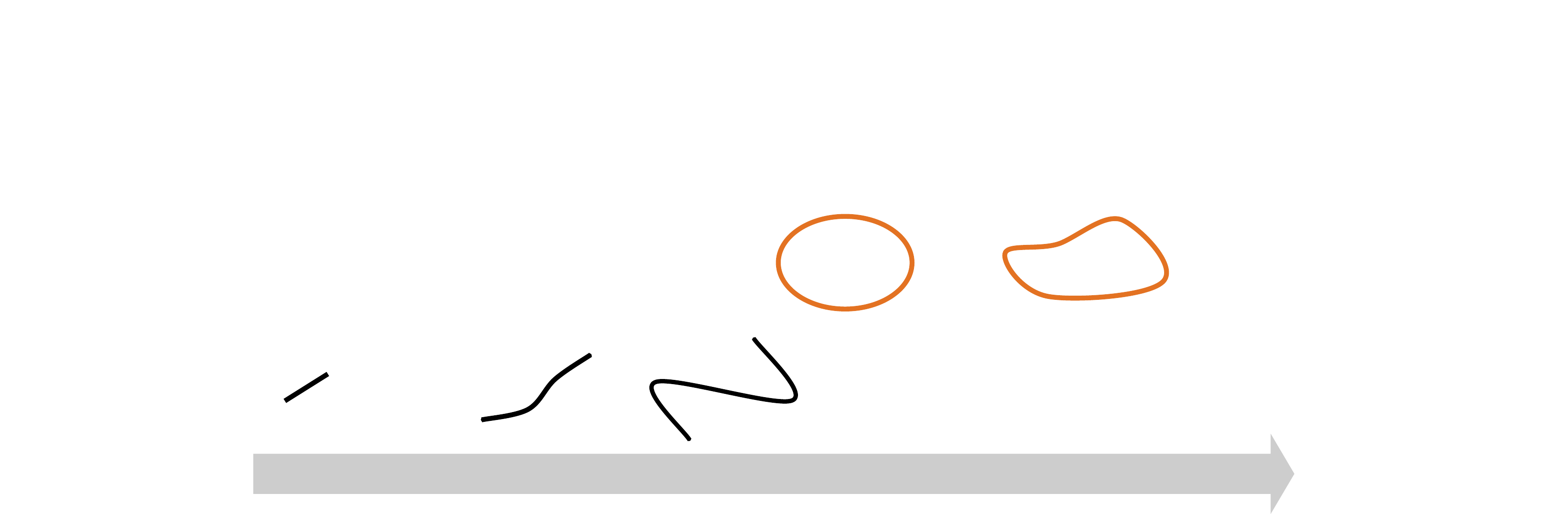}}%
    \put(0.18271263,0.01903061){\color[rgb]{0,0,0}\makebox(0,0)[lt]{\lineheight{1.25}\smash{\begin{tabular}[t]{l}Energy\end{tabular}}}}%
    \put(0,0){\includegraphics[width=\unitlength,page=2]{orbits.pdf}}%
    \put(0.00048537,0.26948193){\color[rgb]{0,0,0}\makebox(0,0)[lt]{\lineheight{1.25}\smash{\begin{tabular}[t]{l}Toroidal Orbit\end{tabular}}}}%
    \put(-0.00241311,0.07305694){\color[rgb]{0,0,0}\makebox(0,0)[lt]{\lineheight{1.25}\smash{\begin{tabular}[t]{l}Brake Orbit\end{tabular}}}}%
    \put(-0.00052914,0.1658662){\color[rgb]{0,0,0}\makebox(0,0)[lt]{\lineheight{1.25}\smash{\begin{tabular}[t]{l}Disk Orbit\end{tabular}}}}%
    \put(0,0){\includegraphics[width=\unitlength,page=3]{orbits.pdf}}%
    \put(0.51374569,-0.00989517){\color[rgb]{0,0,0}\makebox(0,0)[lt]{\lineheight{1.25}\smash{\begin{tabular}[t]{l}$U_{\text{min}}$\end{tabular}}}}%
    \put(0.65943075,-0.01066839){\color[rgb]{0,0,0}\makebox(0,0)[lt]{\lineheight{1.25}\smash{\begin{tabular}[t]{l}$U_{\text{max}}$\end{tabular}}}}%
  \end{picture}%
\endgroup%

  \caption{Overview of types of periodic orbits.}\label{fig:many_orbits}
\end{figure}

Fig.~\ref{fig:many_orbits} summarizes the types of periodic orbits presented in this paper.
One orbit for each type of our example system are also shown on the torus in Fig.~\ref{fig:class_on_torus}.
For very low energies, the linearization around the equilibrium holds as an approximation, and one will have linear modes. 
As energy increases, one observes that the modes begin to bend and we recognize that the linear modes were a particular case of the nonlinear normal modes (brake orbits). 
Indeed, with the continuation method, at least two nonlinear modes can be found for the $2$-DoF system. Our experience so far was that in general, at least $n$ nonlinear modes can be found for $n$-DoF systems. 
Nonlinear normal modes will cease to exist as soon as the total energy exceeds the maximal possible potential energy, i.e.\ if $E> U_{max}(\qq)$. 
In that case, there is no point where all the energy is purely potential and there cannot be points with zero velocity.

Although not predicted so far by the algebraic topology arguments, we have numerically shown that also closed orbits without full rotations exist, which we call disk orbits. 
These might be of particular interest to robotics because these trajectories can be executed also by robots that do not permit endless rotation, which is mostly the case in today's robots. 
Finally, starting at some minimal energy $E>U_{min}(\qq)$, allowing at least on joint to do full turns, closed multi-turn orbits appear. 
For $E> U_{max}(\qq)$, i.e., energies exceeding the maximal potential energy, these toroidal orbits are the dominant periodic behavior. 
There are infinitely many such closed orbits for any robot.

\section{Conclusion}
We hope to have triggered some interest of the robotics community in better understanding the potential benefits of geometric and topological approaches to study the behavior of robot dynamics from a global perspective. Classical robotics control takes a rather local view so far, while global solutions are traditionally the field of motion planning. 
The presented tools might provide a methodical bridge between the two areas. 

Regarding the practical relevance, consider the large variety of periodic orbits we found even in our simplest example.
Complex robots will display even richer behaviors! 
Imagine we can assemble our tasks out of pieces of these orbits - or even better: a task might coincide with a periodic orbit if we design the system properly.
All one needs then is to compensate for friction, stabilize the natural orbits \cite{Bjelonic2022} and possibly develop approaches to shape them to a certain extent, by posture or by control.
By designing and exploiting the intrinsic dynamics of a robotic properly, tasks can be achieved more naturally, more efficiently, and more performantly. 
If we would like a robot to jump like a flea, we should probably not build an elephant but rather something close to a flea.

\section*{Acknowledgments}
We thank Noémie Jaquier and Alexander Dietrich for their feedback!

\printbibliography
\end{document}